# Optimization-based motion planning for autonomous agricultural vehicles turning in constrained headlands

Chen Peng[1] | Peng Wei[2]  | Zhenghao Fei[1] | Yuankai Zhu[3] | Stavros G. Vougioukas[2]

[1]ZJU-Hangzhou Global Scientific and Technological Innovation Center, Zhejiang University, Hangzhou, China

[2]Department of Biological and Agricultural Engineering, University of California, Davis, Davis, California, USA

[3]Department of Mechanical and Aerospace Engineering, University of California, Davis, Davis, California, USA

**Correspondence**
Chen Peng and Zhenghao Fei, ZJU-Hangzhou Global Scientific and Technological Innovation Center, Zhejiang University, Hangzhou, China.
Email: chen.peng@zju.edu.cn and zfei@zju.edu.cn

**Abstract**

Headland maneuvering is a crucial part of the field operations performed by autonomous agricultural vehicles (AAVs). While motion planning for headland turning in open fields has been extensively studied and integrated into commercial autoguidance systems, the existing methods primarily address scenarios with ample headland space and thus may not work in more constrained headland geometries. Commercial orchards often contain narrow and irregularly shaped headlands, which may include static obstacles, rendering the task of planning a smooth and collision-free turning trajectory difficult. To address this challenge, we propose an optimization-based motion planning algorithm for headland turning under geometrical constraints imposed by headland geometry and obstacles. Our method models the headland and the AAV using convex polytopes as geometric primitives, and calculates optimal and collision-free turning trajectories in two stages. In the first stage, a coarse path is generated using either a classical pattern-based turning method or a directional graph-guided hybrid A* algorithm, depending on the complexity of the headland geometry. The second stage refines this coarse path by feeding it into a numerical optimizer, which considers the vehicle's kinematic, control, and collision-avoidance constraints to produce a feasible and smooth trajectory. We demonstrate the effectiveness of our algorithm by comparing it to the classical pattern-based method in various types of headlands. The results show that our optimization-based planner outperforms the classical planner in generating collision-free turning trajectories inside constrained headland spaces. Additionally, the trajectories generated by our planner respect the kinematic and control limits of the vehicle and, hence, are easier for a path-tracking controller to follow. In conclusion, our proposed approach successfully addresses complex motion planning problems in constrained headlands, making it a valuable contribution to the autonomous operation of AAVs, particularly in real-world orchard environments.

**KEYWORDS**

autonomous agriculture vehicle, constrained headland, headland turning, trajectory optimization

Chen Peng and Peng Wei equally contributing authors.

## 1 | INTRODUCTION

Autonomous agricultural vehicles (AAVs), such as field robots and autonomous tractors, are attracting increased attention for modern farming operations. A critical part of autonomous navigation is maneuvering inside the headland space to move to another crop row; this operation involves sharp turns performed by the AAVs at the end of the row. An optimized headland turning maneuver can enhance time and energy efficiency while ensuring operational safety. Motion planning for headland turning relies on a number of factors, including crop type, headland size, and geometry, among others. The primary operating contexts for AAVs fall into two categories: farming in open fields and in orchards.

In open-field farming, crops are typically grown annually or, in some instances, multiple times per year. The cultivation layout can be easily modified within the field's boundaries at the beginning of each cultivation cycle to accommodate the autonomous operation. Open fields usually provide sufficient space for AAVs to perform headland maneuvers, or alternatively, the cultivation layout can be adapted to facilitate these maneuvers. Therefore, research has mainly focused on optimal preplanning of the cultivation layouts, coverage path planning (Jin & Tang, 2010; Mier et al., 2023; Spekken & de Bruin, 2013), and minimizing nonworking distances for vehicles in headlands through global planning of optimal traversal sequences (Bochtis & Vougioukas, 2008). Due to the relatively simple operational context and current technological advancements, Global Navigation Satellite System (GNSS)-based autosteer systems for large vehicles, such as tractors, sprayers, and harvesters, are now commercially available and widely used in open-field farming. Additionally, headland turning maneuvers that do not take into account the obstacles and constrained spaces are also available for commercial autoguidance products (Vougioukas, 2019). Figure 1a illustrates an example of a commercial autosteer agricultural vehicle, and Figure 1b shows a traditional tractor equipped with an autosteer retrofit kit. Both of them can autonomously navigate predefined rows and execute a swath-to-swath turning in open headlands. To the best of our knowledge, these vehicles have nearly replaced human drivers' driving work in open fields.

Unlike open fields, orchards may not always possess spacious, obstacle-free headlands, and expanding orchard headlands by replanting or removing plants can be costly, considering their high value and long lifespan. Most plants grown in orchards are perennial, and the lifespan of an orchard, such as an apple orchard, can range from 20 to 25 years (Lordan et al., 2018). Many existing orchards and crop fields have irregular and limited headland spaces, with some even containing obstacles within the headland. The reason is that these fields were not originally designed for autonomous operations, and orchard owners have sought to maximize land usage by providing minimal headland space. In certain orchards, the headland space is so restricted that even human drivers can barely complete a turn, especially when the vehicle is equipped with a rigidly connected implement (see Figure 2a). Additionally, orchard boundaries may not always have quadrilateral shapes but, instead, irregular shapes, as shown in Figure 2b. The endpoints of tree rows may not lie along a straight line parallel to the boundary (see Figure 2c). In some poorly designed orchards, obstacles such as electricity posts, raised wells, and irrigation infrastructure may be present (see Figure 2d), raising further challenges for autonomous maneuvering following fixed patterns. These factors must be addressed, as they impede the deployment of AAVs in orchards.

This work focuses on the challenging motion planning problem of AAVs turning inside constrained headland spaces, which are common in real-world orchards. We summarize the most common challenges that AAVs may encounter in real-world orchard headlands in Table 1. The first four challenges may not be problematic when sufficient headland space exists. However, combined with the fifth challenge, they can render the motion planning problem challenging. Moreover, even if an orchard has ideal headland conditions in most areas, a small percentage of less-than-ideal headland spaces can hinder the deployment of autonomous operations for AAVs. These challenges may also manifest in open-field farming, where the cultivation layout is not optimized for autonomous navigation, especially when headland space is limited.

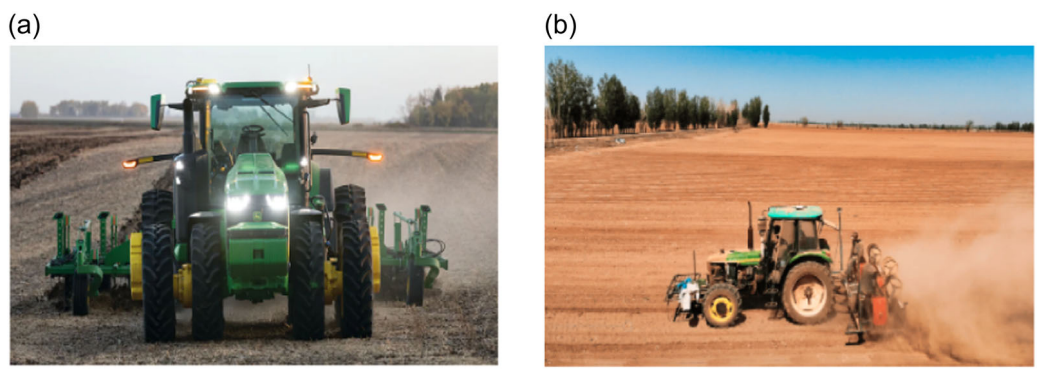

**FIGURE 1** (a) John Deere's autonomous tractor operating in the field (photo courtesy of Deere & Company) and (b) CHCNAV's NX510 autosteer retrofit kit installed on a traditional tractor (photo courtesy of CHC Navigation). [Color figure can be viewed at wileyonlinelibrary.com]

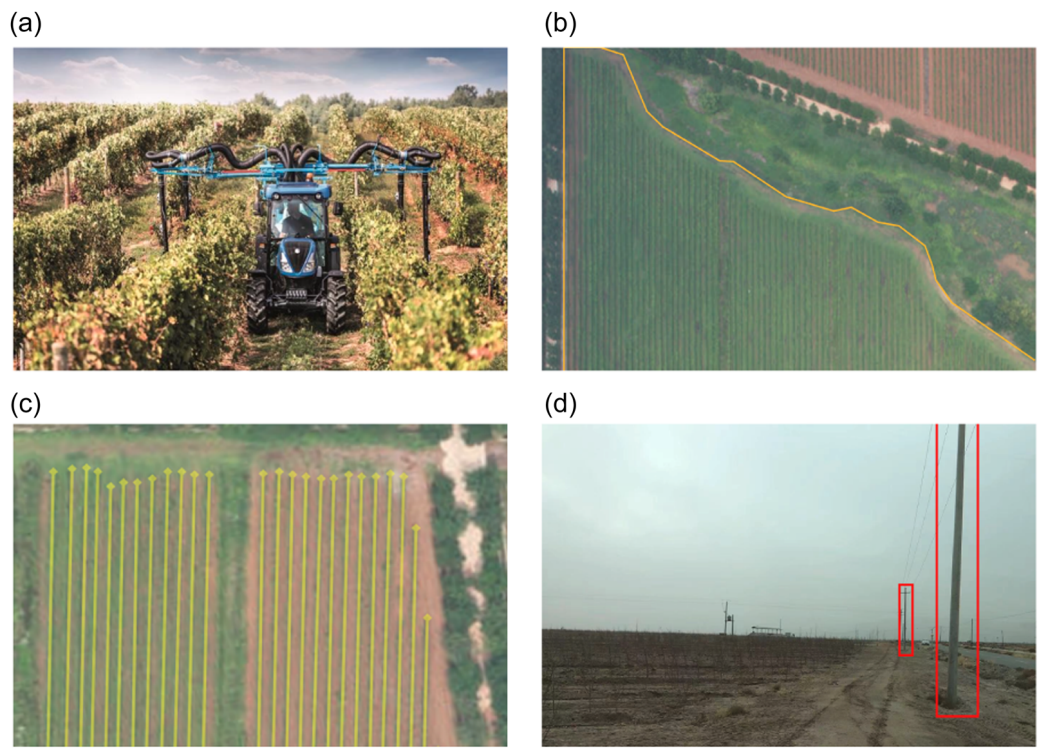

**FIGURE 2** Example of (a) a tractor equipped with a mounted sprayer in a vineyard (photo courtesy of CNH Industrial Company), (b) irregular field boundary, (c) checkered crop row endpoints, and (d) obstacles in headland space. [Color figure can be viewed at wileyonlinelibrary.com]

**TABLE 1** Challenges that autonomous agricultural vehicles encounter in real-world orchard headlands.

| Sources | Challenges |
| --- | --- |
| Tree row | (1) The spacing between tree rows is relatively narrow compared with the vehicle's width. |
| | (2) The endpoints of tree rows may not be collinear. |
| | (3) The vehicle must not run over existing crops in the field. |
| Field boundary | (4) The field boundary can be irregular. |
| | (5) The headland space is small relative to the vehicle's size. |
| Environment | (6) Static obstacles may be present inside the headland area. |

To the best of our knowledge, existing methodologies for AAV motion planning are insufficient for computing effective turning maneuvers in constrained headlands. Thus, developing advanced motion planning algorithms and control strategies to address these challenges is crucial.

In this work, we propose a novel and practical approach for planning turning motions that enable AAVs to navigate in challenging field headland environments. First, we introduce a versatile and generic geometric representation of crop rows, headlands, obstacles, and robots based on collections of convex polygons. Then, we adopt a two-stage planning approach. In the first stage, a collection of fast agriculture-specific row-turning planners and a Hybrid A* planner adapted to crop rows and headlands are used to compute a coarse path. In the second stage, the coarse path warm-starts a numerical optimization algorithm that incorporates suitably represented obstacle avoidance constraints and robot operational constraints to compute the final trajectory. The planner is evaluated in simulations under various field scenarios and in a real-world navigation scenario in a vineyard.

The main contributions of this work are as follows:

1. We present a generic representation of headland space, including irregular field boundaries, irregular crop row ends, static obstacles, and AAVs with or without rigidly attached implements. This representation enhances the accuracy and adaptability of modeling in agricultural settings.
2. We propose a two-stage approach to compute optimal, feasible, and smooth trajectories for AAVs maneuvering in constrained headland spaces. In both stages, we combine established algorithms, incorporate our geometric representation in them, and adapt the algorithms to the specific application scenario.
3. We evaluate the performance of the proposed framework in various simulated headland environments, and compare its success rate and computation time with classic pattern-based methods. We also demonstrate that our framework can be implemented on a real robot to compute complex headland-turning trajectories that can be tracked more accurately than those of classic planners.
4. We open-source our headland-turning planning algorithm in a public repository.[1]

The rest of the paper is structured as follows: In Section 2, we review related work in the literature. Section 3 introduces our proposed optimization-based planning algorithm. Experiments and results are presented in Section 4, followed by a discussion in Section 5. Section 6 concludes the paper.

## 2 | LITERATURE REVIEW

### 2.1 | Dubins and Reeds–Shepp curves

Dubins curves (Dubins, 1957) and Reeds–Shepp curves (Reeds & Shepp, 1990) are commonly used to generate turning paths between rows, such as U-turns and Ω-turns. Custom turning maneuvers such as Switch-back turns (SBTs) and Circle-back turns (Kise et al., 2002; Takai et al., 2011; Torisu et al., 1996; Wang & Noguchi, 2018) are also common in agricultural fields, as illustrated in Figure 3. The lengths of these curves are used as row-to-row transition costs to solve problems related to field coverage path planning and waypoint traversal sequence planning (Bochtis & Vougioukas, 2008; Jin & Tang, 2010; Mier et al., 2023; Spekken & de Bruin, 2013). However, these turning paths exhibit two main drawbacks. The first drawback is that they require the vehicle to always turn at the maximum steering angle and assume that the steering angle can be changed instantaneously, thus ignoring the steering system's dynamic constraints. As a result, the AAV path-tracking controllers cannot track the generated paths adequately, leading to increased tracking errors or even infeasible paths that cause the AAV to move off-track. Recent work by He et al. (2023) proposed dynamic turning path planning for SBTs that alleviates the off-track problem due to side slips in the field. This approach, however, still does not consider the steering dynamics constraints in the planning stage.

The second drawback is that these turning paths are predetermined for each turning pattern and do not explicitly consider the geometries of the field boundaries, crop rows, and any obstacles as hard constraints, which are often irregular and complex, as described in Section 1. Applying such turning paths could easily lead to AAV interference with the boundary, which is often designed to be tight for maximum land usage. Alternatively, modifying the cultivation layout to accommodate irregular field boundaries and crop rows could result in considerable nonplanted areas, thus leading to land waste.

### 2.2 | Continuous curvature (CC) paths

Although Reeds–Shepp paths are easy to generate, they suffer from the issue of discontinuous curvature at the junctions of segments. This results in discontinuous steering angle commands, presenting challenges to the tracking control of nonholonomic vehicles. The vehicle may need to stop and steer to follow the trajectory practically. The problem becomes more serious when subsequent arcs have a minimum turning radius for headland space-saving purposes. To solve this problem, Fraichard and Scheuer (2004) proposed a CC path that smooths the Reeds–Shepp path by ensuring CC. Cariou et al. (2010) presented a motion planner based on primitives connected with clothoid arcs to generate admissible trajectories. Sabelhaus et al. (2013) developed a method for generating CC headland trajectories and investigated the properties of different turning types with CC turns. Backman et al. (2015) introduced a method for generating a smooth turning path with CC and speed profiles, taking into account the vehicle's maximum steering rate and linear acceleration. Plessen and Bemporad (2017) explored three methods for generating smooth reference trajectories and compared their tracking performance using linear time-varying model predictive control. Research on CC path planning has addressed the problem of generating smooth trajectories under steering constraints. However, these methods plan their turning paths, ignoring the constraints imposed by limited headland spaces.

### 2.3 | Optimization-based planning using a simple headland space model

Several research efforts have been made to model and incorporate the headland spatial limits into turning path planning. For instance, Oksanen and Visala (2009) modeled the headland boundaries as parameterized lines and calculated the optimal trajectory by solving an optimal control problem (OCP). Similarly, Tu and Tang (2019) employed a comparable headland space representation and explored a numerical optimization-based method to solve an OCP. They also discovered that the headland boundary angle plays a critical role in the existence of a solution, particularly when the headland width is restricted. These studies

---

[1] https://github.com/AgRoboticsResearch/headland_trajectory_planning

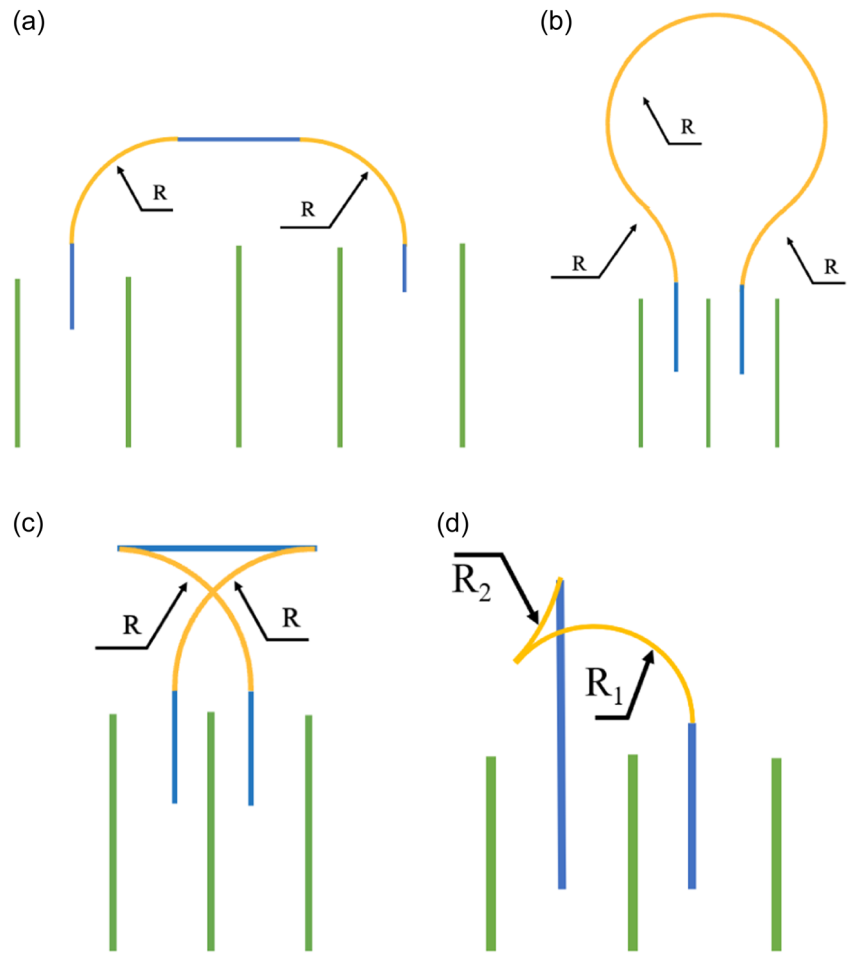

FIGURE 3 Illustrations of (a) *U*-turn, (b) Ω-turn, (c) Switch-back turn, and (d) Circle-back turn. [Color figure can be viewed at wileyonlinelibrary.com]

highlight the growing interest among researchers in identifying feasible turning paths within constrained headland spaces. However, the headland space models in these works are overly simplified and fail to accommodate scenarios where the headland space exhibits irregularities.

## 2.4 | Generic motion planning for agricultural vehicles

Researchers have also explored the generic motion planning problem in agricultural applications, which involves navigating vehicles from a start pose to an end pose while accounting for existing obstacles and environmental structures. Due to the nonconvex and nonlinear nature of this problem, various stochastic optimization techniques, such as genetic algorithms, simulated annealing, and neural networks, have been employed to solve this problem (Ferentinos et al., 2000; Makino et al., 1999; Noguchi & Terao, 1997). However, these methods often suffer from slow convergence and cannot guarantee optimal solutions due to the vast nonconvex search space. The research most closely related to our work in this study is by Vougioukas et al. (2006), who proposed a two-stage optimal motion planner to automatically plan optimal tractor motions. In that work, a sampling-based planner is employed in the first stage to compute a feasible suboptimal trajectory, which is then refined in the second stage under an optimal control framework based on the function-space gradient descent method. However, the computation time for both stages increases dramatically in the presence of obstacles due to the explicit calculation of the minimum distance function at every time step. Additionally, rather than employing constraints, collisions are incorporated as a penalty term into the cost function, which means a collision-free trajectory cannot be guaranteed.

## 3 | METHODOLOGY

This work presents a hierarchical framework to calculate a safe and optimal trajectory for an AAV during headland turning. The framework operates in two stages:

1. *Stage* I: A coarse path is initially computed using a *U*-turn, Ω-turn, or SBT, depending on the minimum turning radius of the vehicle, the number of rows to skip, and the row width. A path is said to be "feasible" if it is collision-free. If the path generated by one of the above turning maneuvers is feasible, it is adopted as the initial guess for the next stage. Otherwise, we apply a kinodynamic path-search algorithm to compute a feasible coarse path, which serves as the initial guess. The details of this stage are presented in Section 3.6.

2. *Stage* II: A numerical optimizer is applied to compute the optimal trajectory, taking the coarse path identified in the first stage as an initial guess. This optimal trajectory minimizes a user-defined cost function, which aims to smooth the trajectory while reducing the total path length and control efforts. The optimizer not only adheres to collision avoidance and vehicle kinematics constraints but also ensures both the state and control variables are within their respective bounds. Details of the numerical optimization are explained in Section 3.5.

A flowchart depicting the proposed framework is shown in Figure 4. The start and end poses are determined by the type of operation. For example, in pruning operations, the start pose is the location where the AAV completes pruning the last tree in one row, and the end pose is where the AAV begins pruning the first tree in the next row.

## 3.1 | System dynamics

Given the state space $\mathcal{X}$ and the control input space $\mathcal{U}$, the vehicle dynamics can be represented in a discrete form as follows:

$$\mathbf{x}_{i+1} = \mathcal{F}(\mathbf{x}_i, \mathbf{u}_i), \quad (1)$$

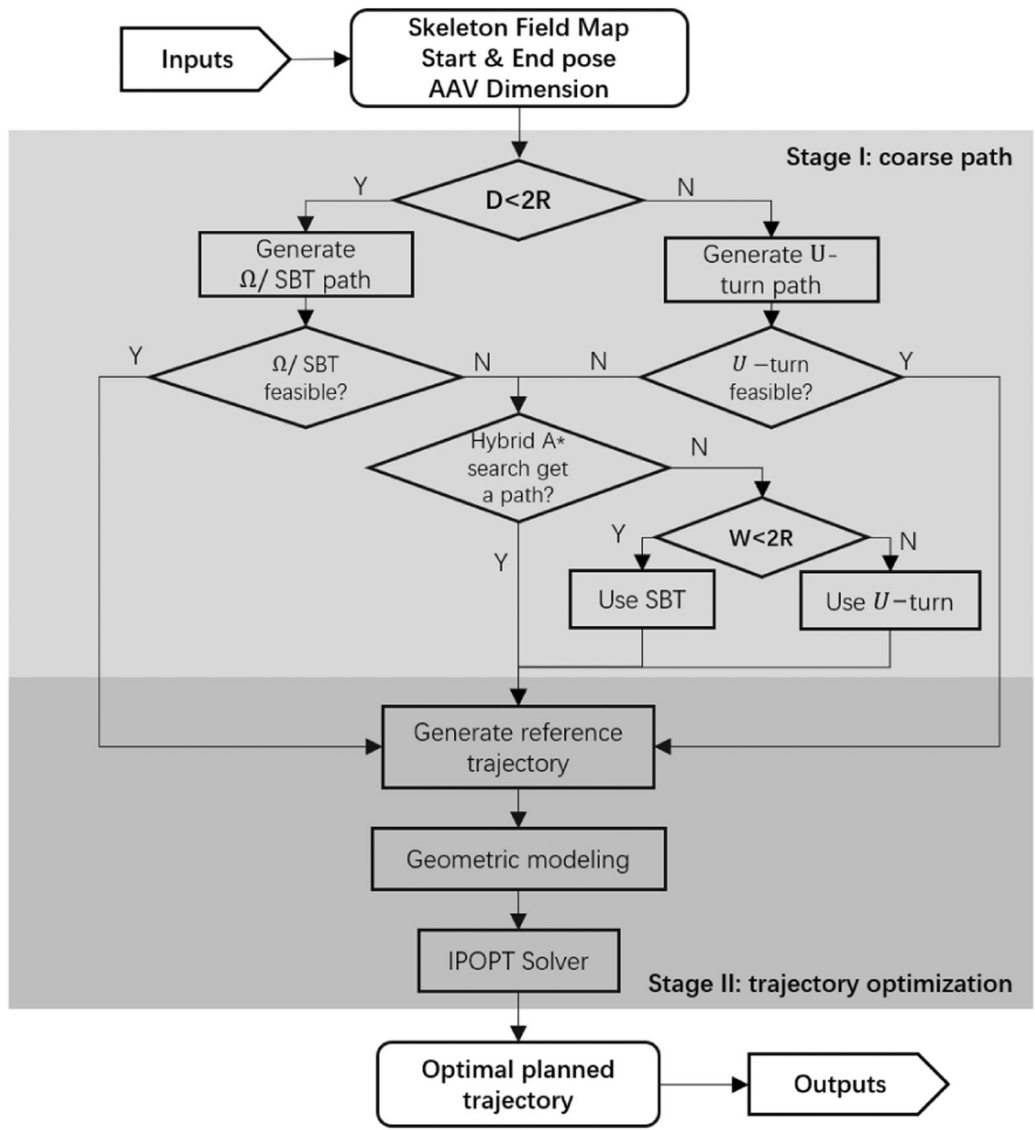

**FIGURE 4** A flowchart of the proposed framework. AAV, autonomous agricultural vehicle; IPOPT, Interior Point Optimizer; SBT, Switch-back turn.

where $\mathbf{x}_i \in \mathcal{X}$ is the vehicle's state vector, $\mathbf{u}_i \in \mathcal{U}$ is the control input vector, $\mathcal{F}: \mathcal{X} \times \mathcal{U} \to \mathcal{X}$ represents the vehicle's motion equation, and $i \in \{0, ..., N\}$ denotes the time step index. The state and control input vectors are subjected to bounded box constraints of the form:

$$\mathbf{x}_{min} \leq \mathbf{x}_i \leq \mathbf{x}_{max}, \quad \mathbf{u}_{min} \leq \mathbf{u}_i \leq \mathbf{u}_{max}. \quad (2)$$

We use a simple kinematic bicycle model to capture the vehicle dynamics in this work. The discretized motion equations are provided in (3).

$$\begin{cases} x_{i+1} = x_i + v_i \cos \theta \Delta T, \\ y_{i+1} = y_i + v_i \sin \theta \Delta T, \\ \theta_{i+1} = \theta_i + v_i \tan \phi_i / L \Delta T, \\ v_{i+1} = v_i + \dot{v}_i \Delta T, \\ \phi_{i+1} = \phi_i + \dot{\phi}_i \Delta T, \end{cases} \quad (3)$$

where $[x_i, y_i]^T$ represents the two-dimensional (2D) position of the vehicle's real-wheel axle midpoint in the global frame. $v_i$ denotes the vehicle's linear velocity along the direction of motion, while $\theta_i$ is the heading angle between $v_i$ and the global $x$-axis. $\phi_i$ is the steering angle of the front wheel. $L$ refers to the wheelbase length, and $\Delta T$ is the discretization time interval. Consequently, we have $\mathbf{x}_i = [x_i, y_i, \theta_i, v_i, \phi_i]^T$. The control input $\mathbf{u}_i = [\dot{v}_i, \dot{\phi}_i]^T$ consists of linear acceleration $\dot{v}_i$ and steering rate $\dot{\phi}_i$. It is worth noting that while the current study uses an Ackermann steering model, the planning algorithm can be applied to alternative steering systems, such as skid steering/crawler, with appropriate modifications to the system dynamics equations.

## 3.2 | Headland environment modeling

In this work, field boundaries, crop rows, and static obstacles within the area of interest, collectively referred to as *obstacles*, are represented by *convex polygons*. The occupied spaces of these polygons are considered nontraversable for the vehicle, and their shapes and locations remain constant over time. These convex polygons are denoted as $\mathbb{O}_1, ..., \mathbb{O}_M \subset \mathbb{R}^2$, where $M$ is the total number of obstacles. Each obstacle polygon can be represented as

$$\mathbb{O}_m = \{\mathbf{z} \in \mathbb{R}^2 : \mathbf{A}_m \mathbf{z} \leq \mathbf{b}_m\}, \quad (4)$$

where $\mathbf{A}_m \in \mathbb{R}^{l \times 2}$, $\mathbf{b}_m \in \mathbb{R}^l$, and $l \geq 3$ denote the number of sides of polygon $m$, assuming a nonempty interior. We will refer to this representation as a *map* throughout the paper. An example of a headland and its map is shown in Figure 5. The endpoints of the lines can be easily acquired using a handheld Real-Time Kinematic-Global Navigation Satellite System (RTK-GNSS) device. For the purposes of this study, we make the following assumptions: (1) we consider a 2D navigation problem, assuming negligible differences in elevation, (2) the map is preconstructed and known to the vehicle, and (3) the vehicle's localization can be obtained from an RTK-GNSS device, with its error being negligible.

Furthermore, we define a *typical field* and a *nontypical field* as follows:

1. A *typical field* is characterized by a headland space defined by two parallel lines. These two lines represent the field boundary and the connection of crop row endpoints, respectively. We assume the crop rows are straight, parallel, and equally spaced apart. Parameters of a typical field include the width of the headland ($D$), the width between two crop rows ($w$), the width of trees ($w_{tree}$), and the angle $\beta$ between the headland line and the row endpoints line. This model is graphically represented in Figure 6. The nontraversable areas formed through such boundaries are represented by quadrilaterals. An illustration of a typical field can be seen in Figure 7a.
2. A *nontypical field*, which more accurately reflects real-world scenarios, is characterized by headland boundaries and row endpoints that cannot be simplified to straight lines. For instance, while the boundary in a typical field can be represented as a simple quadrilateral obstacle, the boundary in a nontypical field is represented using a polyline. Furthermore, the nontraversable areas can be denoted by multiple polygons, as shown in Figure 7b. Crop rows are represented as rectangles in this model.

## 3.3 | Vehicle and implement modeling

AAVs carrying implements can have complex geometries, which, coupled with the presence of arbitrary obstacles and irregular boundaries, render the motion planning for AAVs more challenging than that of problems like autonomous valet parking (Qin et al., 2020), which deal with rectangular cars and perfectly structured parking lots. To accurately model this complexity, we represent both the vehicle and its implements in 2D space as unions of multiple convex polygons. Only implements rigidly attached to the AAV are considered in this study. Figure 8 illustrates two examples of an agricultural vehicle equipped with (a) a pruner and (b) a mower, along with their corresponding polygon representations. Our approach contrasts with the more conservative method of encapsulating the entire robot in a single convex hull, allowing for a more realistic and maneuverable vehicle model and a wider range of scenarios.

Given the current state $\mathbf{x}_i$, the vehicle and any attached implements are modeled as $K$ convex polygons, denoted as $\mathbb{V}_1(\mathbf{x}_i), ..., \mathbb{V}_K(\mathbf{x}_i) \subset \mathbb{R}^2$. The occupied space of $\mathbb{V}_k(\mathbf{x}_i)$ can be alternatively represented by a combination of rotation and translation applied to an initial convex set $\mathbb{B}_k$:

$$\mathbb{V}_k(\mathbf{x}_i) = \mathbf{R}(\mathbf{x}_i) \cdot \mathbb{B}_k + \mathbf{t}(\mathbf{x}_i), \quad \mathbb{B}_k = \{\mathbf{z} \in \mathbb{R}^2 : \mathbf{G}_k \mathbf{z} \leq \mathbf{g}_k\}. \quad (5)$$

In this formula, $\mathbf{R}(\mathbf{x}_i) \subset \mathbb{R}^{2 \times 2}$ is a rotation matrix and $\mathbf{t}(\mathbf{x}_i) \subset \mathbb{R}^2$ is a translation vector. The initial convex set $\mathbb{B}_k$ remains constant, where $\mathbf{G}_k \in \mathbb{R}^{j \times 2}$, $\mathbf{g}_k \in \mathbb{R}^j$, and $j \geq 3$ depends on the number of sides of either the vehicle or its implement, assuming a no-empty interior.

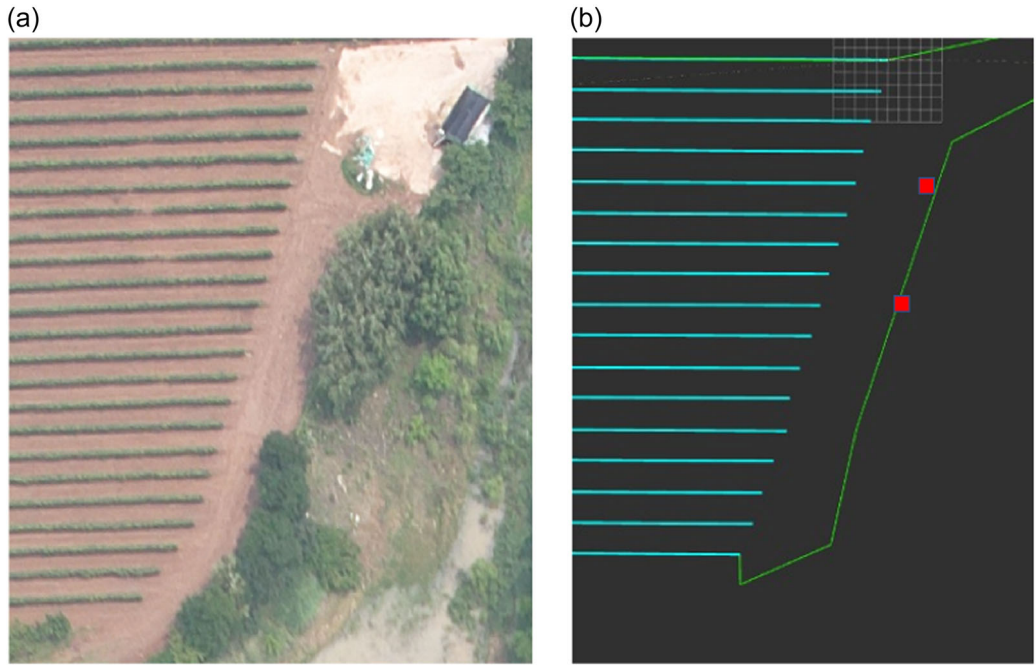

**FIGURE 5** (a) The left figure provides a top-down view of the headland space in a vineyard. (b) The right figure represents the vineyard's map, where the polylines (green) represent field boundaries, the red squares represent the static obstacles, and the rectangles (cyan) represent crop rows. [Color figure can be viewed at wileyonlinelibrary.com]

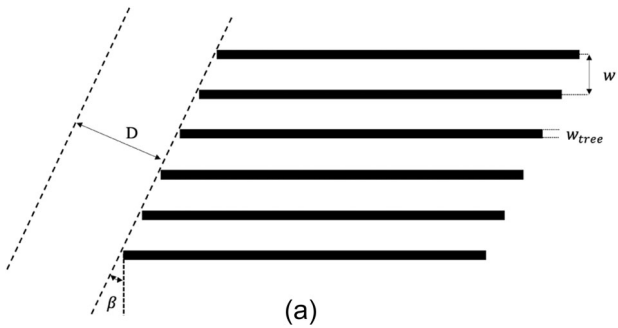

**FIGURE 6** Diagram of a typical field parameterized by the width of the headland ($D$), the width between two crop rows ($w$), the width of trees ($w_{tree}$), and the angle ($\beta$) between the headland line and the crop lines.

From this point forward, we will collectively refer to the vehicle and its implements as the *vehicle*.

## 3.4 | General OCP

The collision-avoidance condition can be formally expressed by (6), which must hold true at every time step. In other words, at each time step $i$, the intersection of the occupied spaces of the vehicle polygons and the obstacle polygons must be empty, meaning that there is no collision between the vehicle and any of the obstacles.

$$\mathbb{O}_m \cap \mathbb{V}_k(\mathbf{x}_i) = \emptyset, \quad \forall\, m \in \{1, ..., M\}, \forall\, k \in \{1, ..., K\}. \tag{6}$$

We formulate the headland motion planning as an OCP. Given a feasible initial state $\mathbf{x}_s$ and terminal state $\mathbf{x}_e$, a set of vehicle polygons $\{\mathbb{V}_1, ..., \mathbb{V}_K\}$ and obstacle polygons $\{\mathbb{O}_1, ..., \mathbb{O}_M\}$, and the total time steps $N$, we need to determine a state sequence $\mathbf{x} = \{\mathbf{x}_0, \mathbf{x}_1, ..., \mathbf{x}_N\}$ and a control input sequence $\mathbf{u} = \{\mathbf{u}_0, \mathbf{u}_1, ..., \mathbf{u}_{N-1}\}$ that minimizes a user-defined cost function $\mathcal{J}(\mathbf{x}, \mathbf{u})$. $N$ is determined by the initial guess trajectory. This cost function optimizes factors, such as total path length, control efforts, and trajectory smoothness. The resulting state sequence must satisfy the collision-avoidance requirement at each time step. The states and control inputs must also adhere to the vehicle's equations of motion and their respective limitations. The representation of this generic OCP is provided as follows:

$$\begin{aligned}
\min_{\mathbf{x},\mathbf{u}} \quad & \mathcal{J}(\mathbf{x}, \mathbf{u}) \\
\text{subject to} \quad & \mathbf{x}_0 = \mathbf{x}_s, \quad \mathbf{x}_N = \mathbf{x}_e, \\
& \mathbf{x}_{i+1} = \mathcal{F}(\mathbf{x}_i, \mathbf{u}_i), \\
& \mathbf{x}_{\min} \leq \mathbf{x}_i \leq \mathbf{x}_{\max}, \\
& \mathbf{u}_{\min} \leq \mathbf{u}_i \leq \mathbf{u}_{\max}, \\
& \mathbb{O}_m \cap \mathbb{V}_k(\mathbf{x}_i) = \emptyset, \\
& \forall\, i \in \{0, ..., N-1\}, \quad \forall\, m \in \{1, ..., M\}, \\
& \forall\, k \in \{1, ..., K\}.
\end{aligned} \tag{7}$$

## 3.5 | Collision-avoidance dual problem

The collision-avoidance constraints in (7) are typically nondifferentiable; hence, the optimization problem is challenging. We adopt the idea from Zhang et al. (2021) to facilitate computation and transform

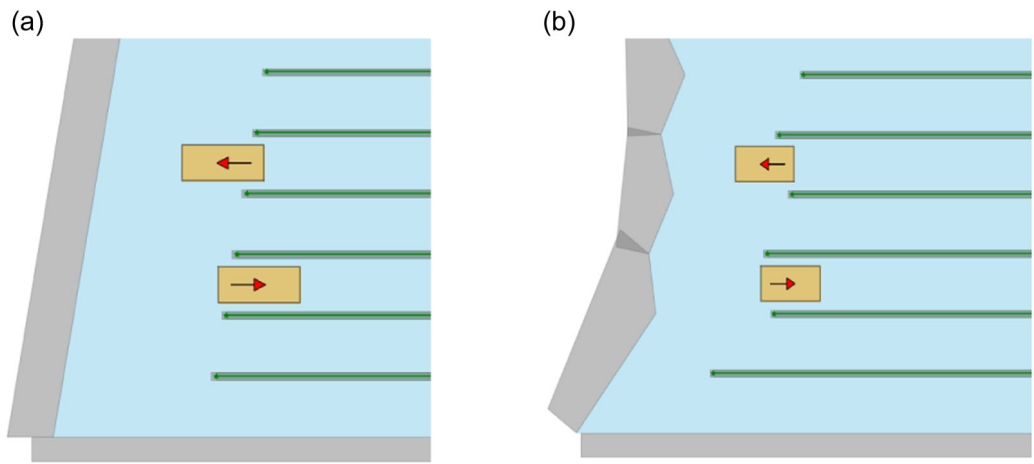

**FIGURE 7** The field boundaries, forming gray nontraversable areas, are depicted as (a) quadrilaterals in a typical field and (b) multiple polygons in a nontypical field. The crop rows are represented by green rectangles. The area traversable by the vehicle is highlighted in cyan, and the start and end poses of the vehicle are also provided. [Color figure can be viewed at wileyonlinelibrary.com]

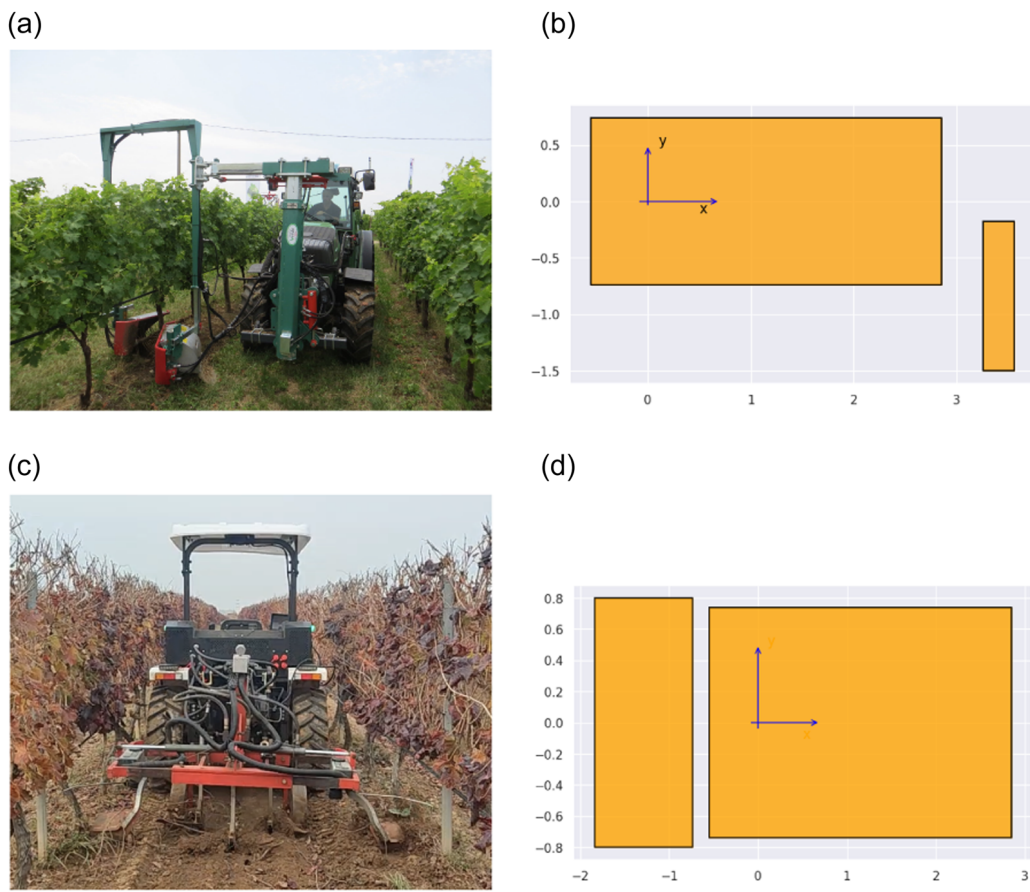

**FIGURE 8** Agricultural vehicles equipped with (a) a summer pruner (photo courtesy of CLEMENS Technologies) and (b) a mower, as seen in the vineyard. Corresponding top-down views of the polygonal models for motion planning applied exclusively for headland turning are shown in (c) and (d), respectively. [Color figure can be viewed at wileyonlinelibrary.com]

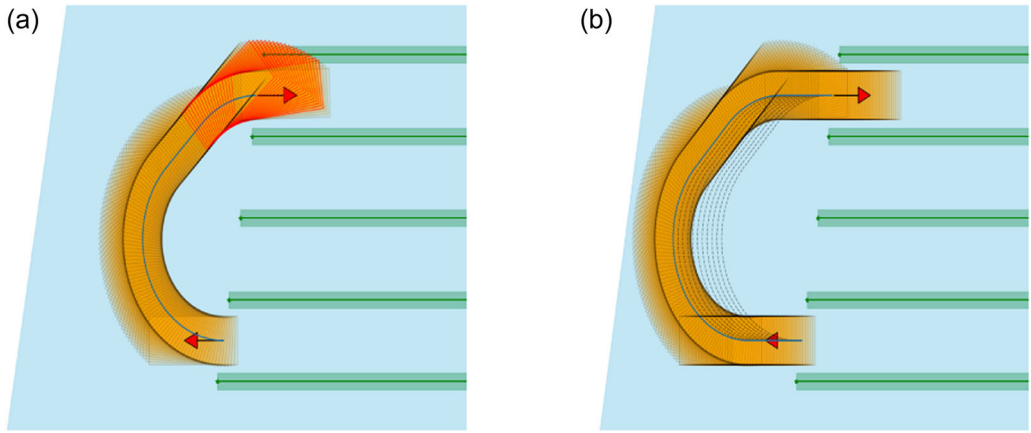

**FIGURE 9** (a) The initial U-turn path that collides with crop rows, and (b) the shifted U-turn path that avoids collisions with crop rows. The black dashed lines represent paths under evaluation, and the solid blue line indicates the path that has been chosen. [Color figure can be viewed at wileyonlinelibrary.com]

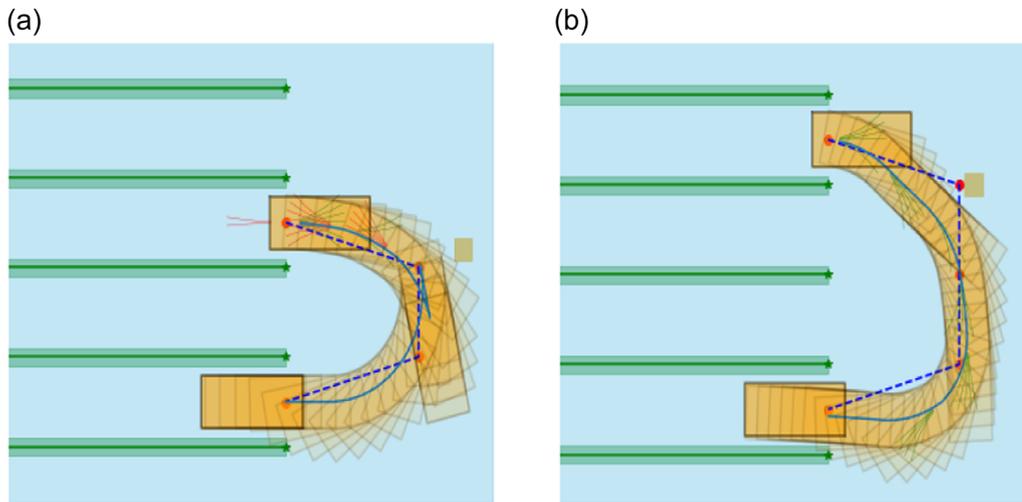

**FIGURE 10** Hybrid A* path based on the directional graph in two different cases: (a) Case I: $d < 2R$ and (b) Case II: $d > 2R$. The green curves represent the expanded nodes' paths in forward motions, while the red curves represent the paths in backward motions. The blue curves represent the searched path. The squares represent static obstacles in the headland. [Color figure can be viewed at wileyonlinelibrary.com]

the nondifferentiable collision-avoidance constraints into smooth, differentiable constraints. Given that $\mathbb{O}_1, ..., \mathbb{O}_M$ and $\mathbb{V}_1(\mathbf{x}_i), ..., \mathbb{V}_K(\mathbf{x}_i)$ are all convex sets, using the signed distance $dist(\mathbb{V}_k(\mathbf{x}_i), \mathbb{O}_m)$ to express the collision-avoidance constraints (Schulman et al., 2014), we can take advantage of the strong duality of convex optimization (Boyd et al., 2004) and reformulate the nondifferentiable collision-avoidance constraints into a set of smoothed convex constraints. The dual problem of $dist(\mathbb{V}_k(\mathbf{x}_i), \mathbb{O}_m) > d_{\min}$, computed at time step $i$, is equivalent to

$$\exists \begin{array}{l} \lambda_i^{m,k} \geq 0: \\ \mu_i^{m,k} \geq 0 \end{array} \begin{cases} -\mathbf{g}_k^T \mu_i^{m,k} + (\mathbf{A}_m \mathbf{t}(\mathbf{x}_i) - \mathbf{b}_m)^T \lambda_i^{m,k} > d_{\min}^{m,k}, \\ \mathbf{G}_k^T \mu_i^{m,k} + \mathbf{R}(\mathbf{x}_i)^T \mathbf{A}_m^T \lambda_i^{m,k} = 0, \\ \|\mathbf{A}_m^T \lambda_i^{m,k}\|_2 \leq 1, \end{cases} \quad (8)$$

where $d_{\min}^{m,k}$ is the desired minimum distance between the two convex sets. $\lambda_i^{m,k}$ and $\mu_i^{m,k}$ are the dual variables associated with the obstacle $\mathbb{O}_m$ and the vehicle part $\mathbb{V}_k(\mathbf{x}_i)$, at time step $i$.

### 3.5.1 | Optimization-based collision-avoidance

By incorporating the duality in (8), the generic OCP outlined in (7) can be reformulated exactly with smooth and differentiable constraints. The effectiveness of this method has been validated in solving the autonomous valet parking problems (Zhang et al., 2021; Zhou et al., 2021). This reformulation, along with our chosen cost function, is presented below:

PENG ET AL. | 11 | WILEY

$$2 \min_{\mathbf{x},\mathbf{u},\lambda,\mu} \sum_{i=0}^{N-1} \mathbf{u}_i^T \mathbf{Q}\mathbf{u}_i + \Delta\mathbf{u}_i^T \mathbf{P}\Delta\mathbf{u}_i + w(v_i \Delta T)^2 \quad (9a)$$

$$\text{subject to } \mathbf{x}_0 = \mathbf{x}_s, \quad \mathbf{x}_N = \mathbf{x}_e, \quad (9b)$$

$$\mathbf{x}_{i+1} = \mathcal{F}(\mathbf{x}_i, \mathbf{u}_i), \quad (9c)$$

$$\mathbf{x}_{\min} \leq \mathbf{x}_i \leq \mathbf{x}_{\max}, \quad (9d)$$

$$\mathbf{u}_{\min} \leq \mathbf{u}_i \leq \mathbf{u}_{\max}, \quad (9e)$$

$$-\mathbf{g}_k^T \boldsymbol{\mu}_i^{m,k} + (\mathbf{A}_m \mathbf{t}(\mathbf{x}_i) - \mathbf{b}_m)^T \boldsymbol{\lambda}_i^{m,k} \geq d_{\min}^{m,k}, \quad (9f)$$

$$\mathbf{G}_k^T \boldsymbol{\mu}_i^{m,k} + \mathbf{R}(\mathbf{x}_i)^T \mathbf{A}_m^T \boldsymbol{\lambda}_i^{m,k} = 0, \quad (9g)$$

$$\|\mathbf{A}_m^T \boldsymbol{\lambda}_i^{m,k}\|_2 \leq 1, \quad (9h)$$

$$\boldsymbol{\lambda}_i^{m,k} \geq 0, \quad \boldsymbol{\mu}_i^{m,k} \geq 0, \quad (9i)$$

$$\forall i \in \{0, ..., N-1\}, \quad \forall k \in \{1, ..., K\},$$

$$\forall m \in \{1, ..., M\}.$$

The cost function (9a) is designed to minimize the control efforts, control input derivatives (i.e., speed and steering jerks), and overall path length. Equation (9b) establishes the initial and terminal condition constraints, while (9c) represents the motion equation constraint for the vehicle. State and control input limitation constraints are encoded in (9d) and (9e), respectively. Collision-avoidance is ensured by (9f), (9g), (9h), and all dual variables must remain positive, as mandated by (9i). In this work, we use the off-the-shelf solver, Interior Point Optimizer (IPOPT), to solve the optimization problem in (9).

## 3.6 | Calculate initial guess

Given the strong nonlinearity of the optimization problem described in (9), both the feasibility and computation time of the solution are greatly dependent on the initial guess, also known as the *warm start*. Identifying an appropriate initial guess is a notable challenge for general optimal motion planning problems. Fortunately, in the context of row-to-row turning in agricultural fields, various turning patterns have been thoroughly investigated and established, as presented in Section 2. In this study, we use an algorithm (Algorithm 1) that chooses one of three classic pattern-based turns to calculate a coarse path: the Ω-turn, the U-turn, or the Switchback turn. The selection among these patterns is made considering the start and end poses, the turning radius, the physical dimension of the vehicle, and the available headland space. We refer to the selection algorithm of the appropriate warm-start turning path as *Approach 1* and describe it next in Section 3.6.1.

If none of these pattern-based paths is feasible, we apply a kinodynamic path-searching algorithm, which operates based on a grid map of the field block, to find a feasible path. This coarse path meets the collision-free requirements and adheres to the kinematic constraints of the AAV, such as minimum curvature constraints. We call it *Approach 2*. We prioritize the pattern-based turning over kinodynamic searching due to its significantly reduced computational time. Furthermore, the pattern-based turning path aligns closer to the optimized path when the headland space is ample and free of static obstacles. However, in complex headland conditions, the search-based path can circumvent obstacles, providing a more viable initial guess for the optimizer.

Given the coarse path and a desired speed profile, a reference trajectory can be obtained. This reference trajectory serves as the initial guess for the optimization problem encapsulated in (9), which enhances computational efficiency and facilitates the discovery of the optimal solution more readily. The procedure of generating the initial guess trajectory can be seen in Stage I of Figure 4. Next, we discuss in detail how to generate a coarse path using the two approaches.

### 3.6.1 | *Approach 1*: Use pattern-based turning

The planner is described using pseudocode in Algorithm 1. The pattern-based turning planner selects one of the three well-established turning maneuvers computed by classic planners: the U-turn, Ω-turn, and SBT (Kise et al., 2002; Sabelhaus et al., 2015). The generated path is a sequence of points separated by a user-defined distance *ds* (set to 0.1 m) and comprises arc and line segments.

The turning pattern is selected based on the vehicle's minimum turning radius (R) and the skip distance (d) between the exit and entry rows. If the distance is smaller than 2R, the robot cannot move on a U-turn to enter the next row. So, the feasibility of an Ω-turn is assessed first. However, Ω-turns require more headland space (see Figure 3b) and are thus more likely to violate the headland boundary. Therefore, the area swept by the robot as it travels on the generated path is checked against the boundary of the headland using a *CollideWithHeadland*() function. If the swept area intersects with the headland boundary, an SBT is generated instead because it uses less space (see Figure 3c). If the skip distance is larger than 2R, there is enough distance to perform a simple U-turn, so a U-turn path is computed.

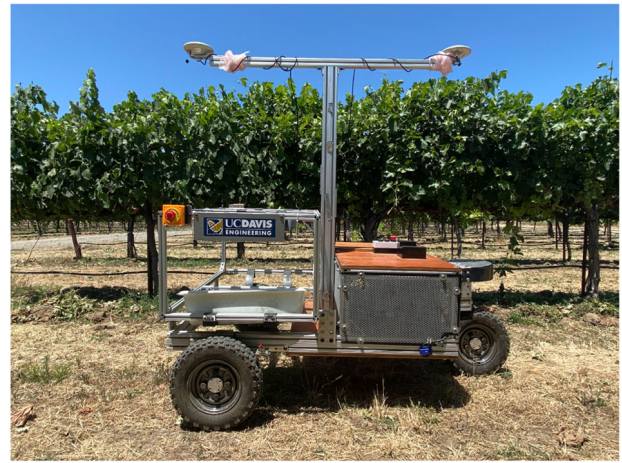

**FIGURE 11** A tricycle mobile robot used in the field experiments. [Color figure can be viewed at wileyonlinelibrary.com]

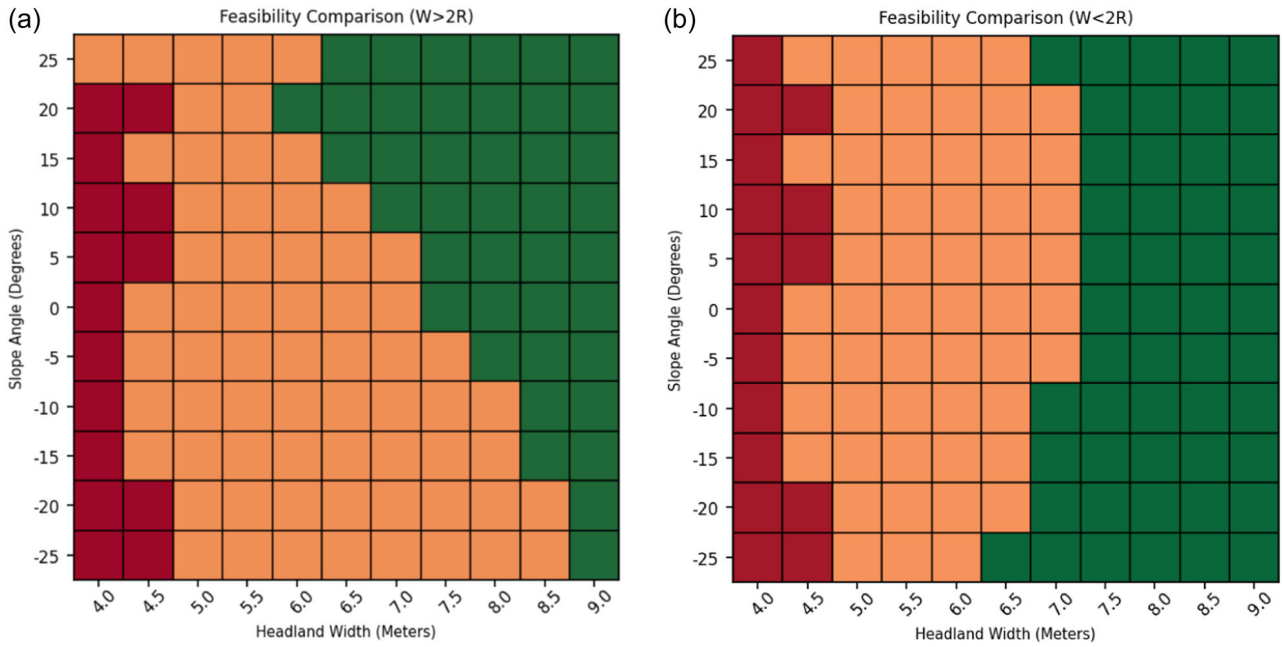

**FIGURE 12** The feasibility results for different $\beta$ and $D$ values in two cases: (a) $W > 2R$ and (b) $W < 2R$. Cell color represents the ability of both planners to compute feasible paths. Red color represents that neither the classic nor our method generated a feasible path. Yellow color represents that the classic planner failed while our method succeeded. Green color represents that both the classic and our method generated a feasible path. [Color figure can be viewed at wileyonlinelibrary.com]

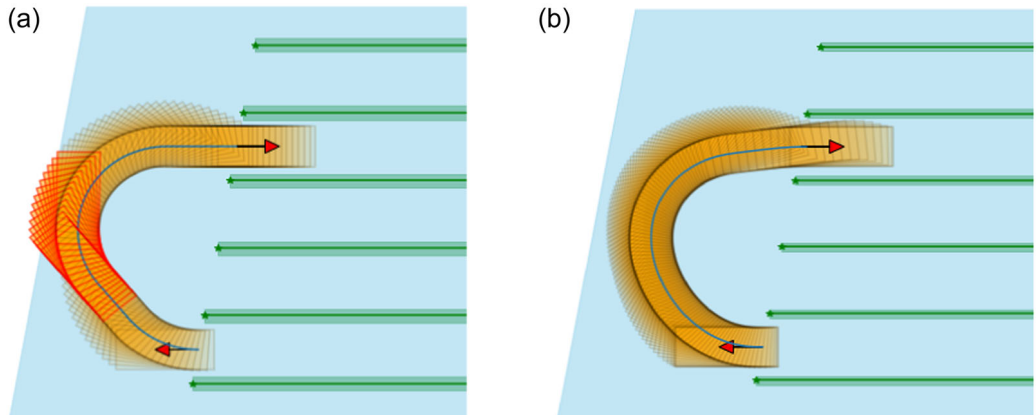

**FIGURE 13** Comparison of path planning results for $\beta = 10°$ and $D = 6.0$ m. (a) Infeasible path generated by the classic planner and (b) feasible path generated by our proposed planner. The blue curve is the planned path, and the vehicle's occupancy at each sampling step is drawn in the figure. Orange shapes denote safe poses, while red shading indicates obstacle-collision instances. [Color figure can be viewed at wileyonlinelibrary.com]

**Algorithm 1** Pattern-based turning planner

**Input:** start pose $\mathbf{x_s} = (x_s, y_s, \theta_s)$, end pose $\mathbf{x_e} = (x_e, y_e, \theta_e)$, turn radius $R$, path resolution $ds$, skeleton map $\mathcal{M}$, shift step $l$.

**Output:** final_path $= [(x_0, y_0, \theta_0), ..., (x_N, y_N, \theta_N)]$.

**Init:** $d = abs(y_s - y_e)$, need_adjust = False

**if** $d < 2R$ **then**

    path = $\Omega$ − Turn($\mathbf{x_s}, \mathbf{x_e}, R, ds$)

    **if** CollideWithHeadland($\mathcal{M}$, path) **then**

        path = SwitchBack − Turn($\mathbf{x_s}, \mathbf{x_e}, R, ds$)

    **end if**

**else**

    path = U − Turn($\mathbf{x_s}, \mathbf{x_e}, R, ds$)

**end if**

**while** CollideWithCrops($\mathcal{M}$, path) **do**

    **for** $i = 0 : length(path) - 1$ **do**

        path$[i][0]$ += $l \cdot \cos(\theta_s)$

        path$[i][1]$ += $l \cdot \sin(\theta_s)$

    **end for**

  need_adjust = True

**end while**

**if** need_adjust **then**

    start_path = GetStraightPath($\mathbf{x_s}$, path$[0]$)

    end_path = GetStraightPath(path$[-1]$, $\mathbf{x_e}$)

    final_path = Concatenate(start_path, path, end_path)

**else**

    final_path = path

**end if**

It is important to note that the algorithm generating the U-Turn(), Ω-Turn(), or SwitchBack-Turn() generates the turning path under a free space assumption, that is, it does not check for collisions against obstacles. Hence, the turning path may cause collisions with the tree/vine rows. This is very often the case when the robot carries an implement, which extends its footprint, thus causing collisions as the robot turns to exit or enter a row. To alleviate this problem, the pattern-based planner checks whether the turning path causes a collision with the crop rows. If it does, the algorithm "shifts" forward to the point where the robot starts to turn to exit its current row by a small distance $l$, to increase the clearance from the crop rows. After the first shift, the algorithm keeps checking for collision with the crop rows and shifting the point forward until the path becomes collision-free. Hence, the original U or Ω or SBT ends up being longer, as it contains a short line segment at its beginning (start_path variable in Algorithm 1) and a short line segment at its end (end_path variable). We refer to this process as *path shifting*. Figure 9 shows an example of path shifting. If the robot traveled on the initial U-turn path (Figure 9a), it would have collided with the top crop row. The black dashed lines in Figure 9b represent the incrementally shifted paths, while the solid blue line depicts the final collision-free path relative to the crop rows.

### 3.6.2 | Approach 2: Use Hybrid A*

The pattern-based output path computed by Algorithm 1 is checked for collisions with obstacles using the CollideWithCrops() function. In situations where the path is not collision-free, a slower yet more powerful Hybrid A* planner is utilized to determine an initial guess. Hybrid A* is a search-based path planner that is capable of generating a smooth path in clustered environments for vehicles with nonholonomic constraints (Dolgov et al., 2010). The search process includes two main steps. First, the node that corresponds to the initial pose is expanded by creating low-cost candidate nodes. The cost is determined using a user-defined heuristic function. Second, an attempt is made to connect all feasible nodes to the goal node using an analytic expansion, such as the Reeds–Shepp or Dubins curves with the given turning radius. The search tree will be expanded until a feasible path is found from the initial node to the node corresponding to the goal pose.

In scenarios such as entering or exiting from parking spaces in parking lots, a directional graph can be constructed from the field map (Qin et al., 2020), and a heuristic cost function can be designed using this graph (Dolgov et al., 2010). We adopted the idea of directional graphs and modified the computation of target areas for agricultural headlands and crop rows. This heuristic function has proven to be substantially more efficient than grid space-based heuristic functions, particularly when the vehicle needs to traverse multiple rows to reach a goal position during headland turning. For detailed proofs and implementations, interested readers are referred to the papers (Dolgov et al., 2010; Qin et al., 2020). Figure 10 presents an example of directional searching in the agricultural field, where path searching is prioritized in the area along the directional graph.

### 3.7 | Trajectory tracking

This study's primary focus is path planning. However, we needed to implement a path tracker to perform real-world field experiments (presented in Section 4.4). The field experiments demonstrated the planner's ability to generate easy-to-track trajectories that adhere to the vehicle's operational constraints without requiring instantaneous velocity and heading changes. After obtaining a smoothed and optimal trajectory, we apply nonlinear model predictive control (NMPC) to achieve precise tracking. NMPC optimizes control inputs over a finite horizon for a nonlinear system while satisfying user-defined constraints. It also exhibits robustness against disturbances. Constrained output path-following using MPC has been discussed in

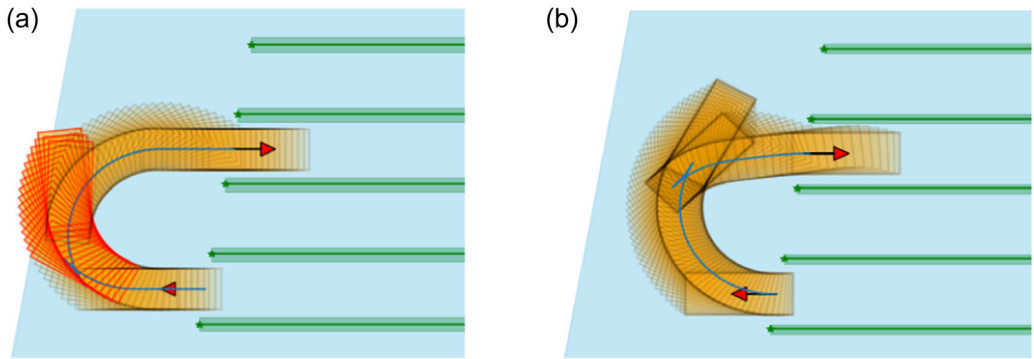

**FIGURE 14** Comparison of path planning results for $\beta = 10°$ and $D = 5.5$ m. (a) Infeasible path generated by the classic planner and (b) feasible path generated by our proposed planner. [Color figure can be viewed at wileyonlinelibrary.com]

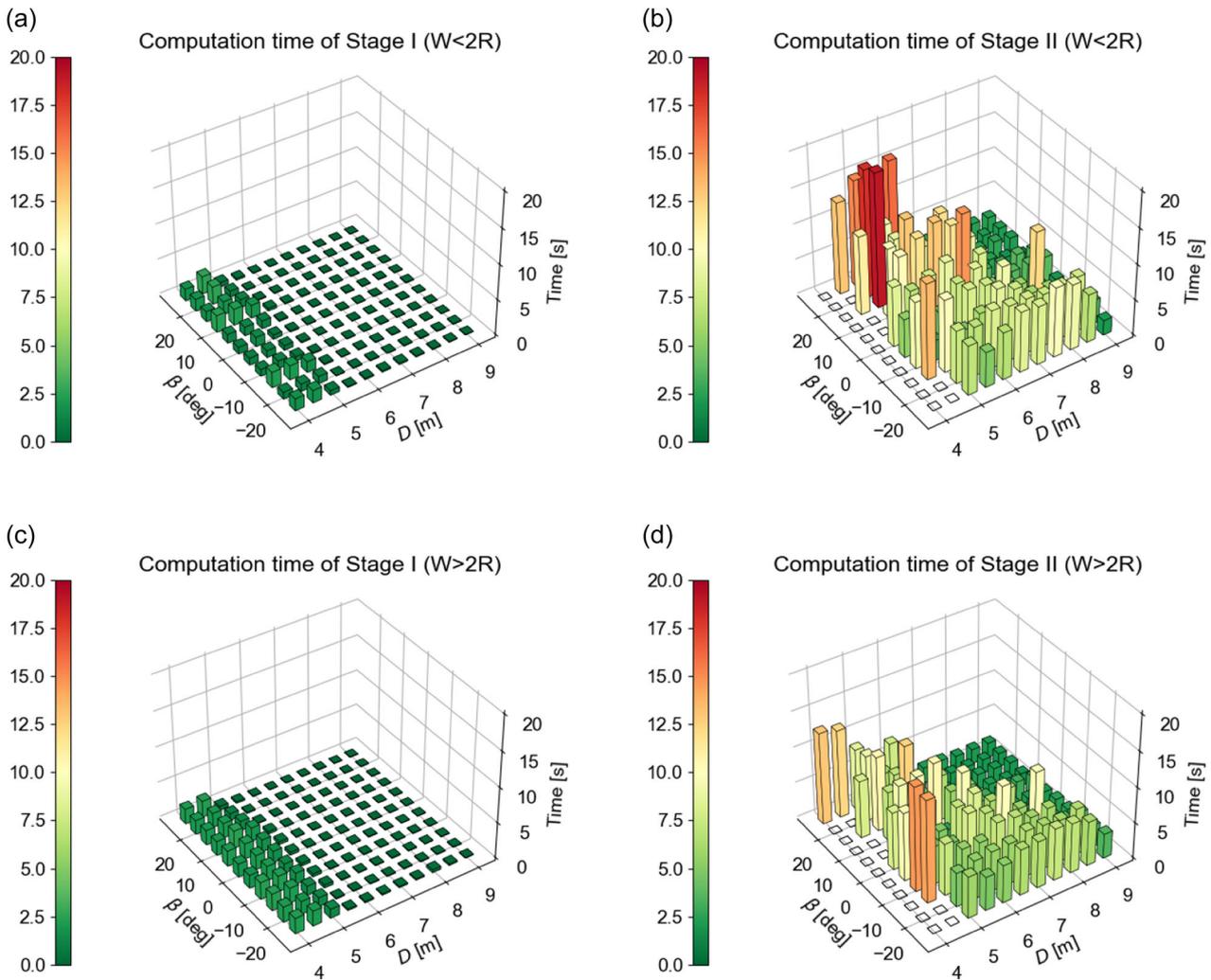

**FIGURE 15** Computation times for Stages I and II of our planner with different $\beta$ and $D$ configurations, where (a) and (b) are the results when $W < 2R$; (c) and (d) are results when $W > 2R$. The times that exceeded the maximum allowed (20 s) have been left blank. [Color figure can be viewed at wileyonlinelibrary.com]

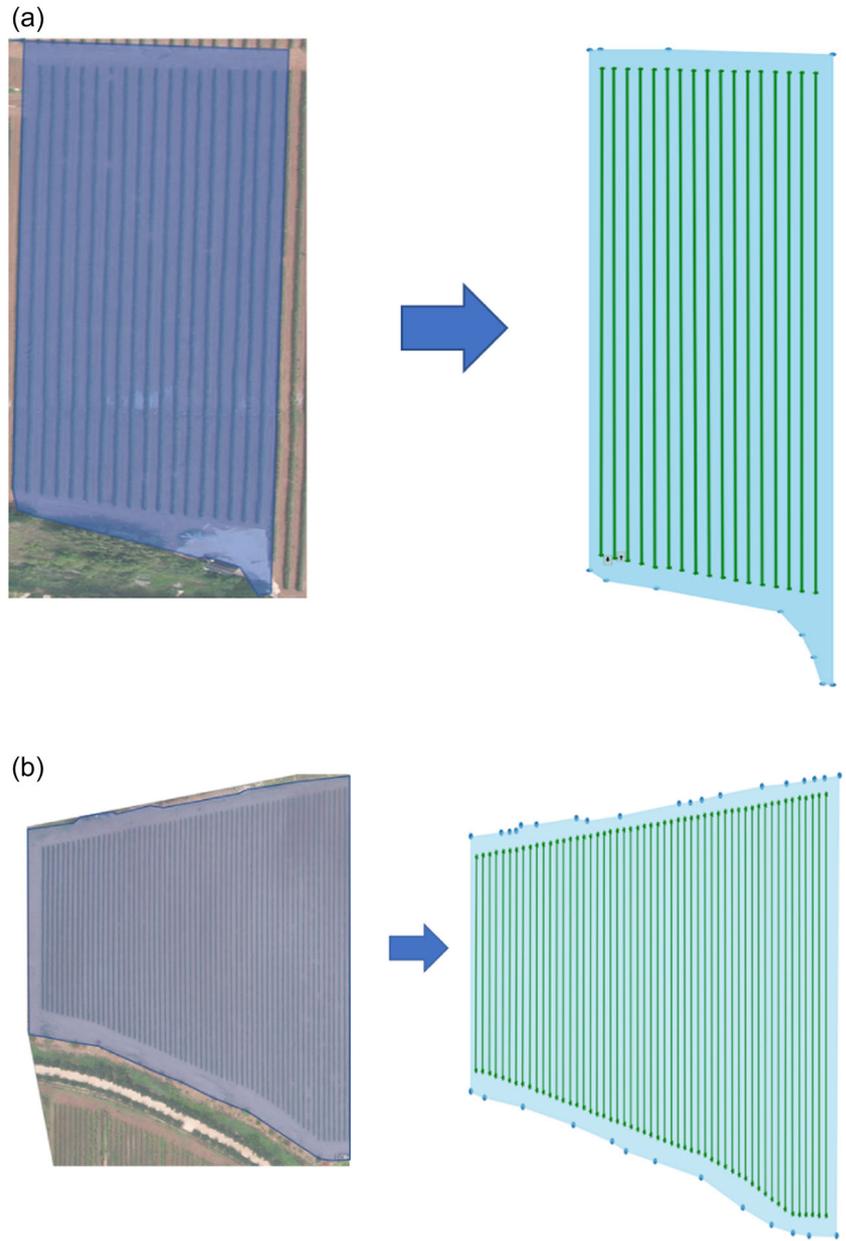

FIGURE 16 Models of two nontypical fields from the real world: (a) Field 1: a less challenging field with 17 rows and (b) Field 2: a more challenging and constrained field with 46 rows. [Color figure can be viewed at wileyonlinelibrary.com]

Faulwasser and Findeisen (2016). By discretizing the system, we can convert the OCP into a nonlinear programming problem, which determines a sequence of control inputs that minimize the deviations from a reference trajectory and reduce control efforts jointly while adhering to constraints on state and control inputs. The formula of our NMPC is provided in (10).

$$\min_{\mathbf{u}} \mathbf{e}_{N_p}^T \mathbf{Q}_e \mathbf{e}_{N_p} + \sum_{i=0}^{N_p-1} \mathbf{e}_i^T \mathbf{Q} \ \mathbf{e}_i + \mathbf{u}_i^T \mathbf{R} \mathbf{u}_i + \Delta \mathbf{u}_i^T \mathbf{P} \Delta \mathbf{u}_i \quad (10a)$$

$$\text{subject to } \mathbf{x}_0 = \mathbf{x}_{init}, \quad (10b)$$

$$\mathbf{x}_{i+1} = \mathcal{F}(\mathbf{x}_i, \mathbf{x}_i), \quad (10c)$$

$$\mathbf{x}_{min} \leq \mathbf{x}_i \leq \mathbf{x}_{max}, \quad (10d)$$

$$\mathbf{u}_{min} \leq \mathbf{u}_i \leq \mathbf{u}_{max}, \quad (10e)$$

$$\Delta \mathbf{u}_{min} \leq \Delta \mathbf{u}_i \leq \Delta \mathbf{u}_{max}. \quad (10f)$$

In the cost function, $\mathbf{e}_i = \mathbf{x}_i - \mathbf{x}_i^{ref}$ refers to the difference between the current state $\mathbf{x}_i$ and the reference trajectory state $\mathbf{x}_i^{ref}$. $\mathbf{Q}$ and $\mathbf{R}$ denote the weighting matrices. To improve the tracking performance, we consider the speed jerk and steering jerk, denoted as $\Delta \mathbf{u}_i$. The corresponding weighting matrix is represented as $\mathbf{P}$ in the cost function. $N_p$ refers to the prediction steps. To make the problem easier to solve, we relax the terminal state constraint and integrate it into the cost function by penalizing its deviation as $\mathbf{e}_N^T \mathbf{Q}_e \mathbf{e}_N$. Here, $\mathbf{Q}_e$ represents the penalty weight, and its magnitude is proportional to $\mathbf{Q}_e = K\mathbf{Q}$, with the value of $K$ chosen empirically. In this work, a state-of-the-art MPC solver (Verschueren et al., 2022) is utilized to tackle the optimization problem in real time.

**TABLE 2** The success rate results from the classic planner and our proposed planner in two nontypical-field blocks.

|  | Total # of turnings | Our planner success rate (%) | Classic planner[a] success rate (%) |
| --- | --- | --- | --- |
| Field 1, w/o implement | 86 | 100 | 73 |
| Field 1, w/mower | 86 | 100 | 51 |
| Field 1, w/pruner | 86 | 100 | 32 |
| Field 2, w/o implement | 289 | 90 | 9 |
| Field 2, w/mower | 289 | 69 | 7 |
| Field 2, w/pruner | 289 | 47 | 3 |

[a]Success with classic planners is achieved when either a $U$ or $\Omega$ turn (if $W > 2R$), or either a Switch-back or Circle-back turn (if $W < 2R$), can generate a collision-free path.
Abbreviations: w, with; w/o, without.

## 4 | EXPERIMENTS AND RESULTS

We conducted four sets of experiments (three sets in simulation and one set in a vineyard) to evaluate the performance of our proposed methodology. The results from our optimization-based motion planner were compared against those obtained from classic pattern-based planners (Jin & Tang, 2010; Kise et al., 2002; Wang & Noguchi, 2018) across a range of headland configurations. More specifically, when the skip distance between rows was adequate ($d > 2R$), $U$ and $\Omega$ turns were computed; otherwise, Switch-back and Circle-back turns were tried (see Figure 3). Since the classic pattern-based turns are computed without considering the presence of the crop rows, path shifting was applied to all of them; otherwise, most pattern-based turns would result in collisions.

We broadly categorized our experiments into four distinct scenarios:

(1) *Typical field*: The first set of simulation experiments benchmarked our planner against classic pattern-based planners in typical fields. The definition of a typical field is provided in Section 3.2.
(2) *Nontypical field*: The second set of simulation experiments compared the performance of our planner with that of classical pattern-based planners in two nontypical fields. The definition of a nontypical field is outlined in Section 3.2.
(3) *Complex headland*: In the third set of simulation experiments, we assessed our planner's performance in a complex headland with narrow spaces and static obstacles, which poses a significant challenge to the classic pattern-based planner. This experiment demonstrated the capability of our planner to navigate more constrained headland spaces.

(4) *Real-world field*: This set of field experiments evaluated our planner's performance in a headland on a campus vineyard. Obstacles were added inside the headland space, and the headland boundary was restricted to recreate the complex headland geometries from the third set of experiments in the real world. These experiments assessed the performance of the entire autonomous navigation system, incorporating both turning trajectory planning and tracking.

In scenarios 1 and 2, we model the vehicle geometry as a rectangle with dimensions of $3.8 \times 1.5$ m. The wheelbase is assumed to be $L = 1.9$ m. The steering angle is limited between $\delta \in [-0.6, 0.6]$ rad, the vehicle velocity is limited between $v \in [-1, 2]$ m/s, the linear acceleration is between $a \in [-0.6, 0.6]$ m/s$^2$ and the steering rate is constrained $\dot{\delta} \in [-0.7, 0.7]$ rad/s. The above parameters correspond to a medium-sized orchard tractor. The sampling time of the trajectory planner is selected as 0.2 s. We set the width of crop row $w = 2.5$ m, and the width of the vines is assumed to be $w_{tree} = 0.4$ m, which is typical for vineyards.

In scenarios 3 and 4, we implemented our planning and control algorithms on a tricycle robot (shown in Figure 11) that operated on a campus vineyard. This robot represents a class of smaller agricultural robots and was used primarily because it has been developed by our group; hence, we had full access to its sensing, control, and navigation software stack. Furthermore, similarly to the orchard tractor model used in scenarios 1 and 2, the tricycle robot's kinematics are described by the bicycle model. The robot has a wheelbase of 1.3 m and dimensions of $1.8 \times 1.2$ m. The steering angle is constrained between $\delta \in [-0.6, 0.6]$ rad and the velocity is constrained between $v \in [-1, 1]$ m/s. The linear acceleration is constrained to $a \in [-0.6, 0.6]$ m/s$^2$, and the steering rate is limited to $\dot{\delta} \in [-0.7, 0.7]$ rad/s. Computations for scenarios 1, 2, and 3 were performed on a laptop with Intel i7-12700H central processing unit (CPU) and 16 GB memory. For the field experiment in scenario 4, the NMPC control algorithm was executed on an Intel NUC computer equipped with an Intel i5-7300 CPU and 32 GB memory.

### 4.1 | Scenario I: Typical field

We first compared the performance of our planner against the classic pattern-based planners in typical fields. Both planners were used to plan a path enabling the vehicle to transition from one row to another. The turning width, denoted as $W = n \times w$, is determined by the number of rows skipped, $n$, and the crop width, $w$. Two different headland turning cases were considered: (1) $W \geq 2R$ and (2) $W < 2R$. In case 1, the vehicle was directed to turn into the third row on the right, allowing for a feasible forward-only path. Conversely, in case 2, the vehicle was guided to turn into the adjacent row on the right, necessitating both forward and backward maneuvers.

In both cases 1 and 2, we varied the headland width $D$ from 4 to 10 m with a resolution of 0.25 m and adjusted $\beta$ from $-25°$ to $25°$ with a resolution of $2.5°$ to explore the space of typical-field headlands. The evaluation metric was a Boolean variable set to "1" if the planner

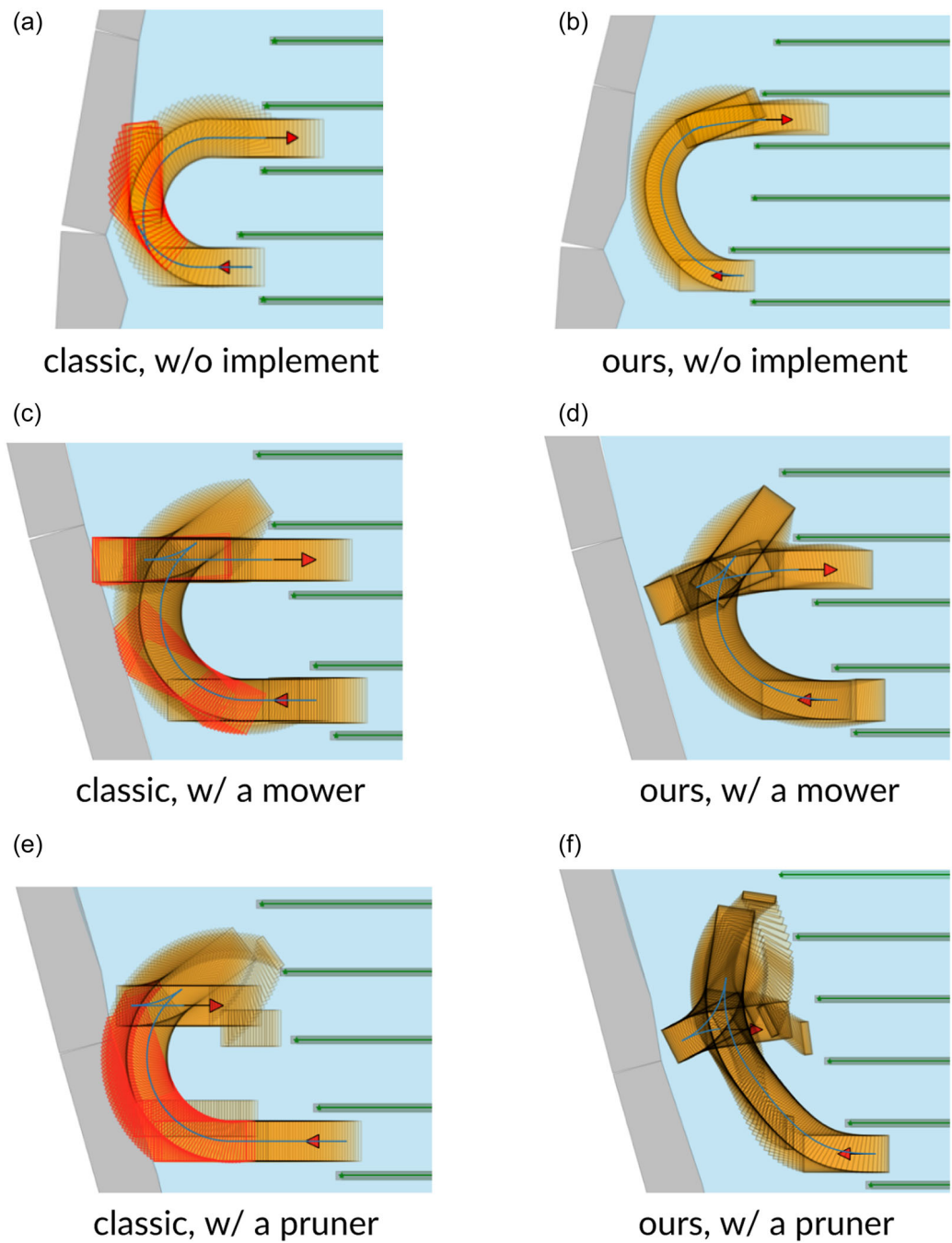

**FIGURE 17** Our planner can effectively utilize irregular space of the headland to maneuver while the classic planner fails to provide a safe path. (a), (c), and (e) depict the path generated by the classic planner without an implement, with a mower, and with a pruner, respectively; (b) without an implement, (d) with a mower, and (f) with a pruner illustrate the path derived from our planner under the same condition, showcasing its effectiveness in each case. [Color figure can be viewed at wileyonlinelibrary.com]

generated a *feasible* path; otherwise, it was set to "0." A generated path was considered feasible if it met all collision-avoidance constraints within a maximum computation time of 20 s.

The results of the first case can be viewed in Figure 12a, and those of the second case can be seen in Figure 12b. Each cell in those figures corresponds to a headland configuration ($\beta$ and $D$ pair). The color of each cell indicates the ability of both planners to compute feasible paths. Red cells represent that neither the classic nor our method can generate a feasible path. Yellow cells represent that the classic planner fails to generate a feasible path while our method succeeds. Green cells represent that both the classic and our methods succeed in generating a feasible path. One should emphasize that our planner's solution is a comprehensive trajectory that respects the vehicle's operational constraints (e.g., maximum velocity and steering angle limits). In contrast, the classic pattern-based planner only provides a rudimentary path and does not

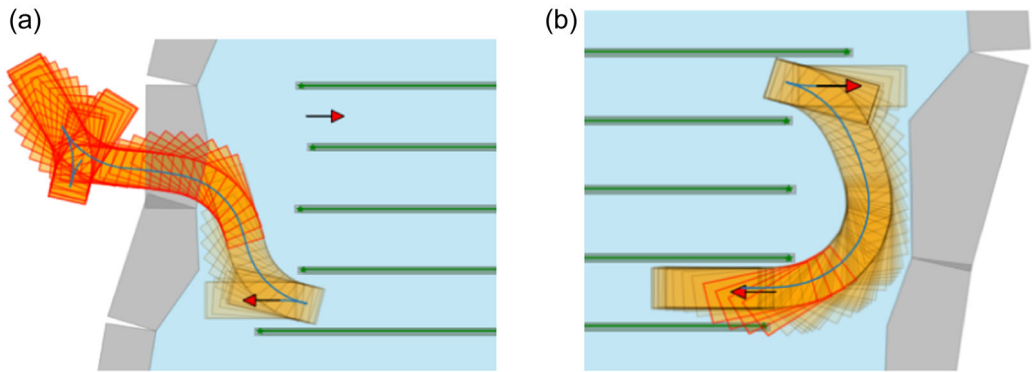

**FIGURE 18** Two failure cases in field 2 with our planner due to excessively narrow headland spaces: (a) AAV went out of the boundary; (b) AAV collided with crop rows. The displayed paths are the last solutions generated by the solver before timeout.

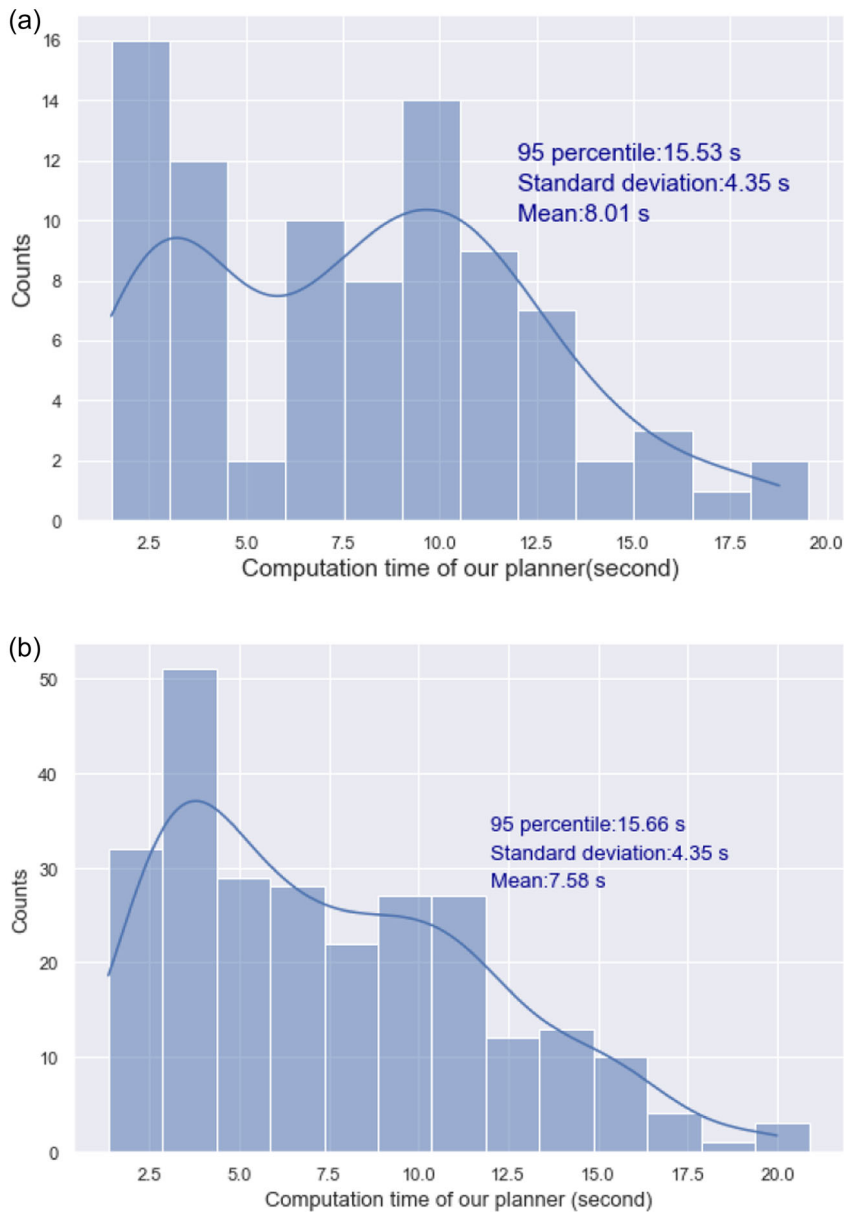

**FIGURE 19** Computation time distributions for our planner are examined in (a) field 1 and (b) field 2 under Scenario II for the vehicle operating without any implements. [Color figure can be viewed at wileyonlinelibrary.com]

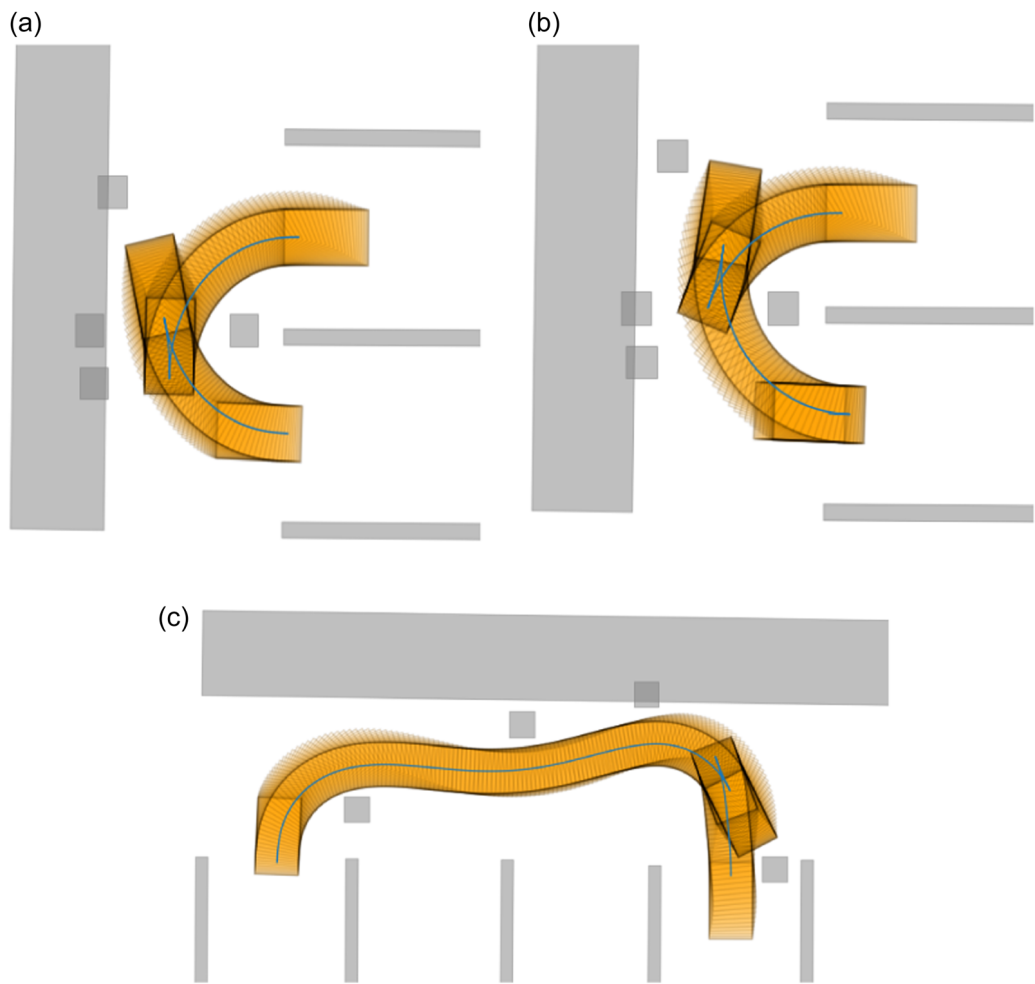

**FIGURE 20** (a), (b), and (c) are three test cases used to evaluate our planner's performance in complex headlands with narrow spaces and static obstacles.

account for the vehicle's constraints except for the minimal turning radius; hence, the path may not be practical for the vehicle to follow.

Figures 13 and 14 provide illustrative examples of the planned paths, which can help readers intuitively understand the difference between our planner's result and the classic planner's result. Figure 13 shows the planned path under case 1 using the forward-only setting for both planners, with an angle $\beta = 10°$ and headland width $D = 6$ m. Figure 14 presents the planned path under case 2, with both planners utilizing Switch-back paths, featuring an angle $\beta = 10°$ and headland width $D = 5.5$ m. In both cases, our planner can determine a solution that successfully completes the turn, while the classic planner's solution is infeasible due to collisions.

We selected computation time as a metric to represent the difficulty of each turn-planning scenario. Consequently, a longer computation time typically signifies a more challenging problem. The total computation time of our planner can be segmented into Stage I (coarse path) time and Stage II (trajectory optimization) time, as illustrated in Figure 15. The characteristics of the headland significantly influence the computational duration of our planner. For Stage I, as shown in Figures 15a,c, the computation becomes noticeably quicker with increasing row width. In most cases where the row width exceeds 5 m, the pattern-based method can generate a feasible coarse path. However, as the space in the headland becomes more constricted, the computation time for the initial guess starts to increase due to the involvement of the Hybrid A* search. For Stage II, as depicted in Figures 15b,d, it is evident that the optimization process becomes faster as the row width $W$ increases. Furthermore, the optimizer's computational duration extends when the value of $\beta$ is negative. This scenario is similar to the "nose-in" parking challenge often encountered in street parking situations. When $\beta < 0$, it represents maneuvering a car into the row in the appropriate direction. In contrast, when $\beta > 0$, it signifies the reversal of the car's direction, a considerably more complex maneuver, as reflected in the planner's computation time.

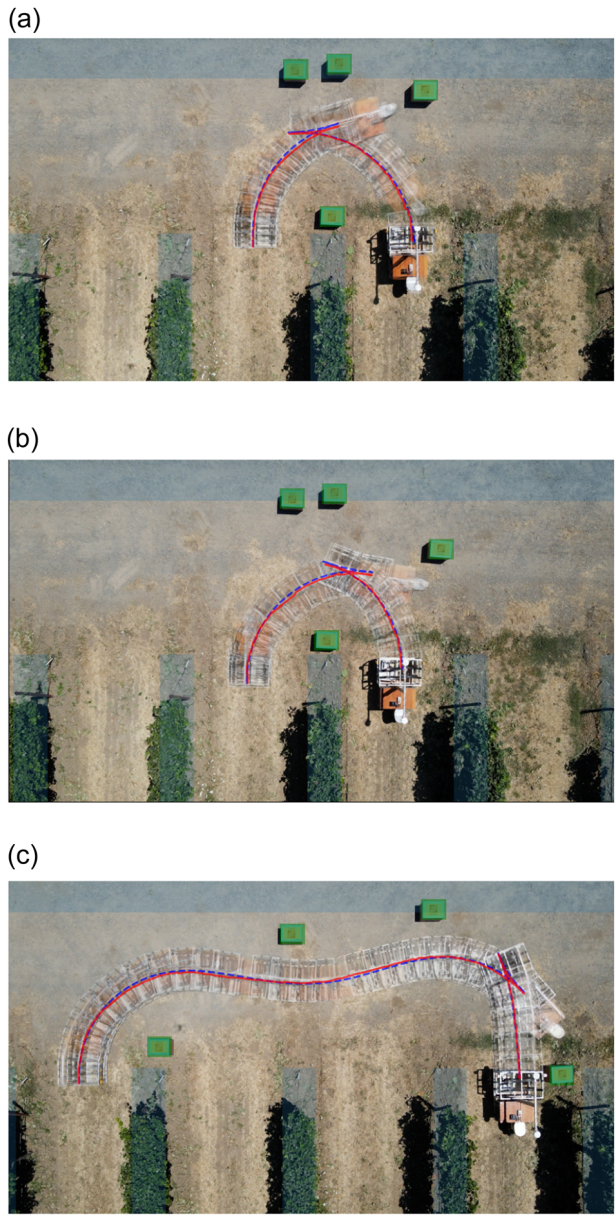

**FIGURE 21** (a), (b), and (c) are three test cases used to evaluate the performance of our method on a mobile robot in a real-world field. The blue dashed line is the planned path and the red solid line is the actual path of the robot. The green rectangles visualize the obstacles' locations in the headland. The field boundary and crop rows' locations are projected to represent their relative positions graphically.

## 4.2 | Scenario II: Nontypical field

The typical-field model often understates the complexity of headland spaces. Headlands can be irregular in real-world fields, and the field boundaries and crop row endpoints do not necessarily align in straight lines. We designed a second experiment to assess our planner's performance in two nontypical fields with irregular field boundaries and crop row ends (see Figure 16). The field maps were derived using an RTK- GNSS receiver. Field 1 consists of 17 rows in total. Field 2 contains 46 rows in total and offers more challenge for the vehicle to make a turn in the headland compared with field 1.

We conducted comprehensive evaluations of headland turnings from each row to its neighboring four rows, comparing the performance of our planner against the classic pattern-based planners. In this study, we selected the success rate as our evaluation metric. Classic planning succeeds if any one of the classic planners can generate a collision-free path. The success rate results for both methods are presented in Table 2. These results indicate that our planner consistently outperforms classic planners in terms of success rate across both fields. When operating without an implement, our planner achieved 100% success in field 1 and 90% in field 2. In contrast, classic planning achieved success rates of only 73% and 9%, respectively. Even with implements that lower the success rates for both approaches, our planner continues to outperform the classic planner and achieves a substantially higher success rate. Figure 17 illustrates a nontypical-field scenario in which classic pattern-based planners failed (Figure 17a,c,e), whereas our planner successfully navigated the irregular headland space in all three cases: without an implement, with a mower, and with a pruner (Figure 17b,d,f). These examples demonstrate that our planner can effectively maneuver inside irregular headland spaces, whereas the classic planner fails to provide a collision-free path. However, it is important to acknowledge that our planner may struggle to find a viable solution in extremely challenging situations, especially when the computation time is limited (20 s). Figure 18 presents two instances where our method failed in excessively narrow headland spaces within the given time limit.

The distributions of the computation time with our planner in both fields when operating without any implements are shown in Figure 19. In field 1, the 95th percentile time is 15.52 s, with a standard deviation of 4.35 s and a mean computation time of 8.01 s. When the pruning implement was added, more complex maneuvering was required, and more polygons—and optimization constraints—were added, thus causing the 95th percentile time, standard deviation, and mean computation time to increase to 17.89, 3.11, and 10.63 s, respectively. Similarly, when the mowing implement was added, the 95th percentile time, standard deviation, and mean computation time were 16.44, 4.21, and 9.26 s, respectively. In field 2, the 95th percentile time was 15.66 s, the standard deviation remained at 4.35 s, and the mean computation time was 7.58 s. When the pruner was added, the 95th percentile time, standard deviation, and mean computation time were 17.31, 3.96, and 10.42 s, respectively. The 95th percentile time, standard deviation, and mean computation time were 16.85, 4.37, and 9.40 s when the mower was added.

## 4.3 | Scenario III: Complex headland

Next, we assessed our planner's performance in complex headlands with narrow spaces (4 m width) and static obstacles. This evaluation aims to showcase our algorithm's ability to handle more intricate headland conditions, particularly where the turning space is limited, and additional collision-avoidance requirements must be enforced to avoid existing obstacles. It is important to note that classic pattern-based planners cannot handle such complex scenarios.

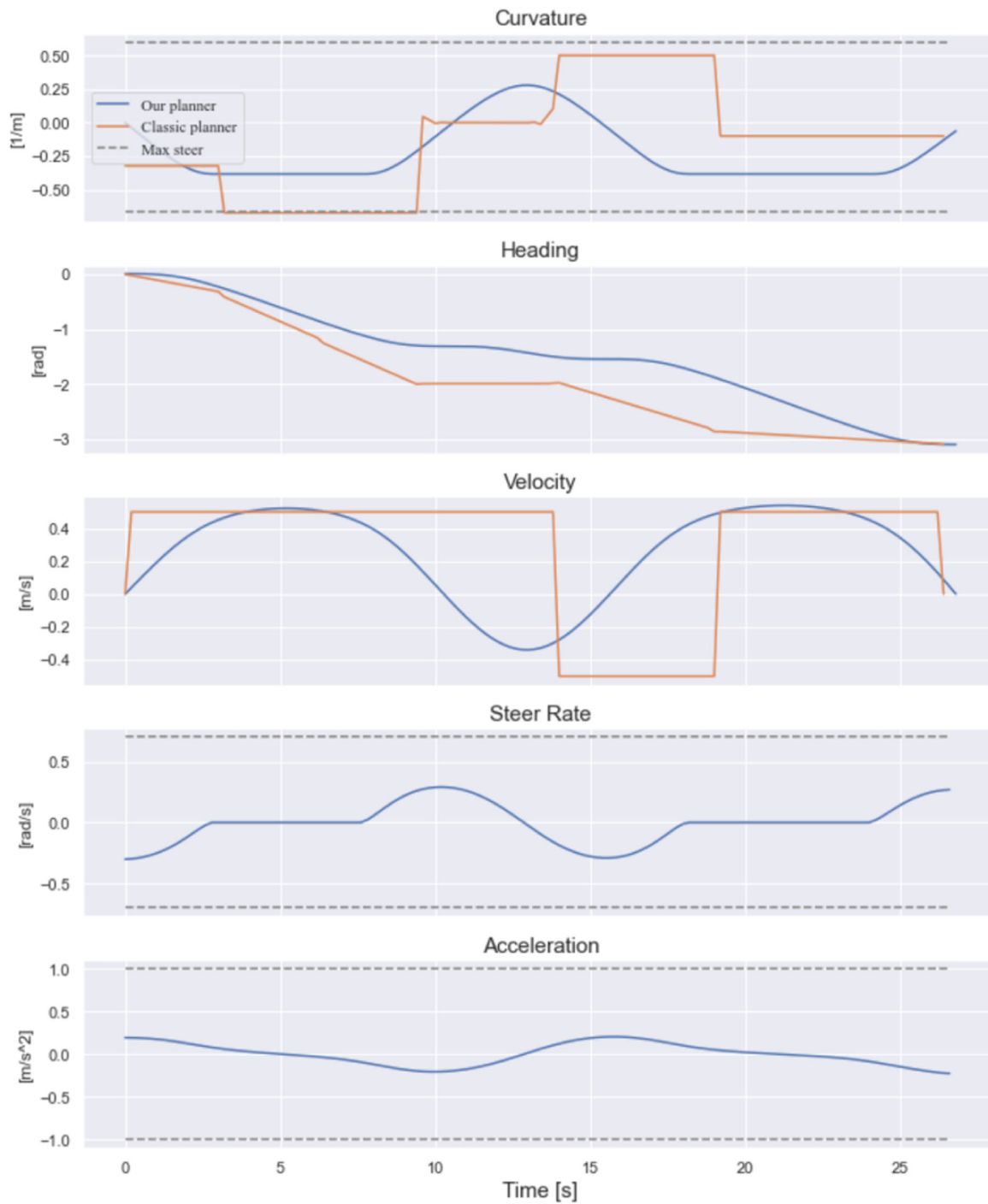

**FIGURE 22** Comparison of the trajectory characteristics generated by the coarse path planner and our planner. The coarse path planner does not output steer rate and acceleration. [Color figure can be viewed at wileyonlinelibrary.com]

We examined three cases, as illustrated in Figure 20. The planning results demonstrate that our planner can effectively generate collision-free paths within these constrained spaces, successfully navigating around static obstacles near the field boundary and crop rows. Since the classic planner is not designed to handle static obstacles, it was not applied in this scenario. The

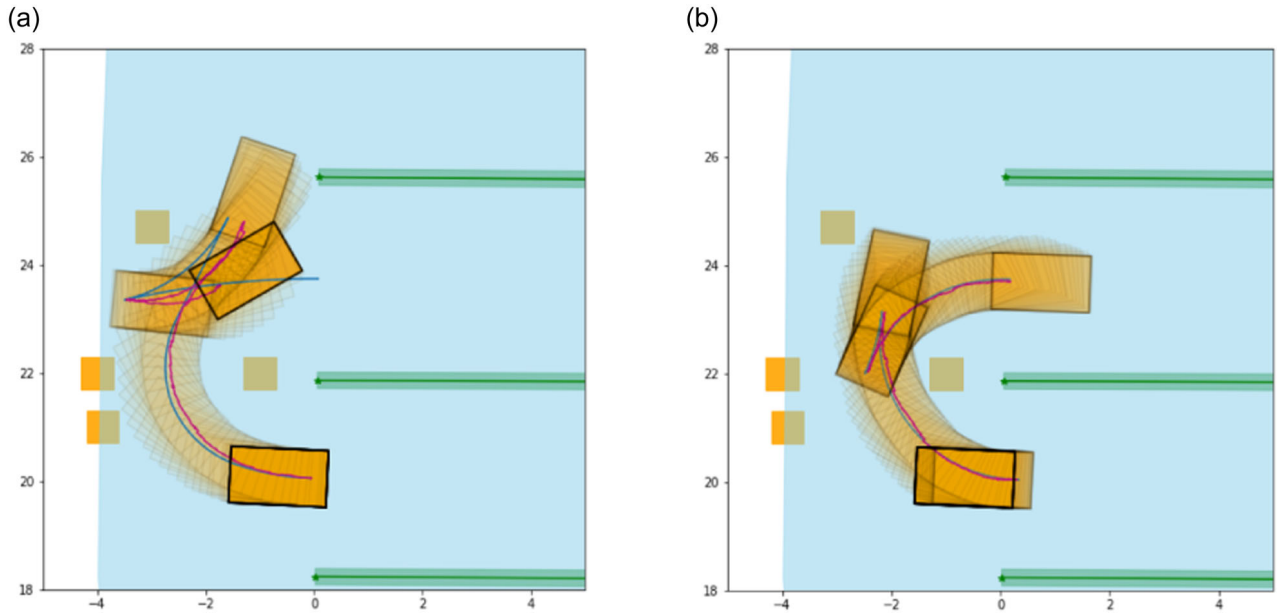

**FIGURE 23** (a) The planned path from the Hybrid A* planner and its tracking results and (b) the planned path from our planner and its tracking results. The blue and red curves show the planned and actual trajectories, respectively. [Color figure can be viewed at wileyonlinelibrary.com]

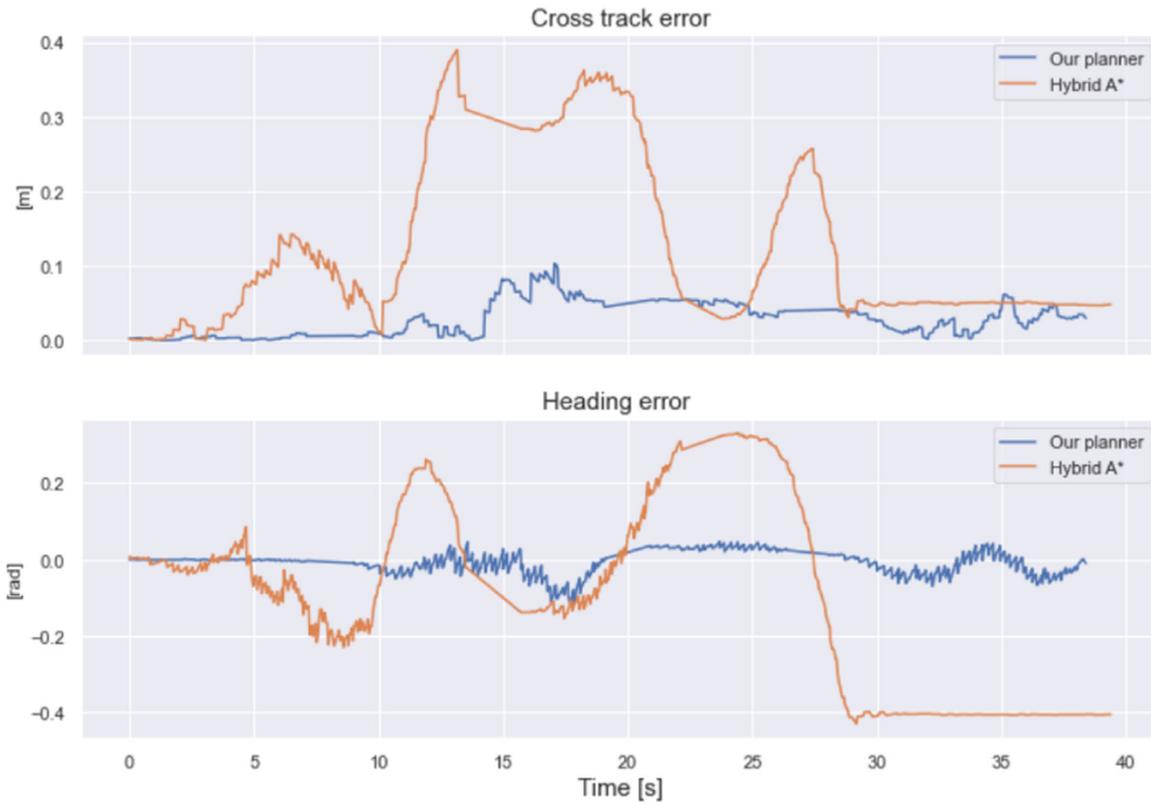

**FIGURE 24** The cross-tracking errors and heading errors of the trajectory generated from the Hybrid A* planner (orange) and our planner (blue). [Color figure can be viewed at wileyonlinelibrary.com]

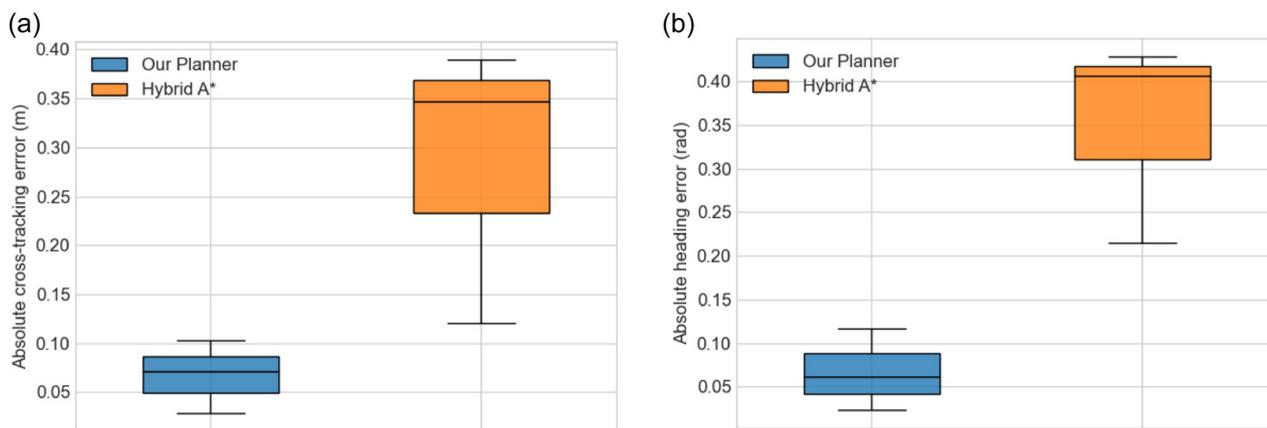

**FIGURE 25** Comparison of (a) the absolute cross-tracking error statistics and (b) the absolute heading error statistics between the Hybrid A* planner and our planner. [Color figure can be viewed at wileyonlinelibrary.com]

computation times of these three cases are 8.2, 9.5, and 12.3 s, respectively.

## 4.4 | Scenario IV: Real-world field

In the final evaluation stage, we tested our entire system, incorporating the planning and tracking modules in a real-world field setting. The purpose of this experiment was to evaluate the performance of our autonomous navigation system in a realistic environment. While the primary focus of this study is path planning, testing in a real-world environment is important to demonstrate the practical effectiveness of our planner. The experiment was conducted in a vineyard located in Davis, CA, using a tricycle mobile robot, as shown in Figure 11. The vineyard's geometry was accurately mapped using an RTK-GNSS receiver. To recreate the complexity of the headland conditions from the previous experiment (refer to Section 4.3), we added artificial obstacles and constrained boundaries in the headland space. All static obstacles' locations were known to the planner (obstacle avoidance was not meant to be dynamic). The robot localization was achieved through an onboard RTK-GPS, fused with an IMU and wheel odometer, running an Extended Kalman Filter in the background. Our planner, deployed on the robot, aimed to compute feasible maneuver trajectories given the field map, the static obstacles' locations, and the start and end poses. Once our planner computed a trajectory, the tracking was executed by an NMPC, which was described in Section 3.7. The NMPC ran at a rate of 30 Hz with a prediction horizon of 2 s. We manually tuned the weight matrices to values of $\mathbf{Q} = [1, 1, 2, 5, 0]$, $\mathbf{R} = [0.5, 0.5]$, $\mathbf{P} = [0.005, 0.005]$, and $\mathbf{Q}_e = 20\mathbf{Q}$, which empirically yielded good tracking performance.

We present the planning and tracking results of three cases in Figure 21. The blue dashed line represents the planned path, while the red solid lines show the actual path. We positioned paper boxes within the headland to help visualize the obstacles' locations. Moreover, we projected the locations of the field boundaries and crop rows used in the planning onto the photograph to graph their relative positions. The results clearly demonstrate our planner's ability to plan a feasible and smooth trajectory within a real-world setting that can be tracked by a real robot. The tracking error across all three cases remained below 0.1 m. For the complete field experiments, please refer to the supplementary video.[2]

We also illustrated the efficacy of our planner in generating a smooth and optimal trajectory in Figure 22. This figure presents the characteristics of two trajectories in our algorithm: (a) the reference (initial guess) trajectory derived from the coarse path using a classic pattern-based planner after Stage I, and (b) the optimized trajectory computed by our planner after Stage II. For simplicity, we will refer to the former as the coarse path planner. These characteristics include curvature, heading, linear velocity, steering rate, and linear acceleration. The resultant trajectory from our planner is depicted in blue, while the orange lines represent the values derived from the coarse path planner. The black dashed lines represent the bounds for each variable (if such bounds exist). The results show that the output trajectory from our planner is significantly smoother than the one from the coarse path planner. More importantly, the planned trajectory adheres to the vehicle's operational constraints without requiring instantaneous value changes. This compliance with constraints makes the trajectory easier to track, demonstrating the advantage of our planning method over classic pattern-based methods.

Additionally, we evaluated the tracking performance of the trajectories generated by (a) a Hybrid A* planner and (b) our planner in the presence of static obstacles within the headland space. As depicted in Figure 23, the mobile robot demonstrated superior efficiency in tracking the smoothed trajectory produced by our planner in contrast to the trajectory created by Hybrid A*.

---

[2]https://www.youtube.com/watch?v=sf0uDFwpSfo

Additionally, according to Figure 24, both the cross-tracking errors and heading errors from the Hybrid A* were noticeably larger than those encountered when tracking the smoothed trajectory generated by our planner. The mean, max, and 95 percentile values of the absolute cross-tracking errors for our planner are 0.028, 0.103, and 0.071 m, while the Hybrid A* are 0.120, 0.390, and 0.347 m. The mean, max, and 95 percentile values of the absolute heading errors for our planner are 0.023, 0.117, and 0.061 rads, while the Hybrid A* are 0.215, 0.429, and 0.406 rads. The results can be visualized in Figure 25. This reduction in errors demonstrates the practicality and effectiveness of our planner's generated trajectory followed by a real robot.

## 5 | DISCUSSION

An important feature of our planner is its ability to generate near-optimal maneuvers when a solution is available. Our algorithm consistently aims for the best possible trajectories, whereas human drivers might not always reach the same level of accuracy. Factors like lack of experience or fatigue can lead human drivers to make suboptimal decisions, especially during tasks like row switching, which result in nonproductive time. Thus, by optimizing such maneuvers, our planner has the potential to significantly increase operational efficiency.

Our planner also has limitations that we plan to address in future work. Currently, our method is limited to rigidly connected implements. One future research direction is to expand this method to be compatible with articulated implements, which are common in agricultural equipment. Also, there are cases, such as the extremely narrow headlands discussed in Section 4.2, where an experienced driver with a typical medium-sized tractor can successfully perform an intricate maneuver, whereas our planner may take a long time to generate a solution or fail to find one. One way to look at this shortcoming is to recognize that our planner's difficulty or failure suggests that the farmer should consider physically improving the headland conditions (e.g., widening the headland) and consequently simplify the required maneuvering. However, it is also true that our planner's performance relies on the quality of the warm-starting solution and the efficiency of the solver. More work is needed in both areas.

The performance of trajectory planning and tracking systems relies on the robot's available computational power and the performance of its actuation systems. Our NMPC currently runs at a rate of 30 Hz using a prediction horizon of 2 s. An extended prediction horizon can enhance the tracking performance as the robot's commanded speed increases. However, a longer horizon would impose a greater computational load on the onboard computer. It is also worth noting that significant time delays in the speed and steering systems can compromise the tracking performance, underscoring the need for timely responses from the actuators. Even though our algorithm was field-tested on a tricycle robot with limited computing power and slow dynamics, our algorithm can be applied to a broad spectrum of modern agricultural robots and electric tractors equipped with enhanced design and actuation systems and computational resources.

## 6 | CONCLUSION

In conclusion, this paper presented a novel methodology for optimization-based motion planning in constrained headlands. Our approach consistently surpassed the classic pattern-based methods by generating feasible and optimal trajectories. We validated our method's applicability and robustness under various field configurations. Our planner consistently outperformed pattern-based planners in typical-field scenarios, achieving feasible and smooth trajectories in a larger range of headland configurations. We also conducted a comparative analysis in two nontypical orchard field blocks, demonstrating the success rates of our planner exceeded those of the pattern-based method in each respective field, with or without implements attached. Additionally, we have demonstrated its effectiveness in planning a safe and optimal trajectory in headlands with narrow spaces and static obstacles. Real-world experiments on an autonomous robot in a vineyard further showcased the applicability of our method to complex environments. In short, this work offers an advanced and practical approach to complex motion planning in constrained headlands, making it a valuable contribution to the autonomous operation of AAVs, particularly within orchards.

### ACKNOWLEDGMENTS
The authors wish to acknowledge that this work was completed without external funding.

### CONFLICT OF INTEREST STATEMENT
The authors declare no conflict of interest.

### DATA AVAILABILITY STATEMENT
The data that support the findings of this study are available from the corresponding author upon reasonable request.

### ORCID
Peng Wei  http://orcid.org/0000-0003-0491-2480
Stavros G. Vougioukas  http://orcid.org/0000-0003-2758-8900

## SUPPORTING INFORMATION

Additional supporting information can be found online in the Supporting Information section at the end of this article.